\pdfoutput=1
\documentclass[twoside]{article}
\usepackage[accepted]{aistats2024}
\usepackage{mathtools}
\usepackage{amsthm}
\usepackage{microtype}
\usepackage{graphicx}
\usepackage{subfigure}
\usepackage{color}
\usepackage{microtype}
\usepackage{graphicx}
\usepackage{subfigure}
\usepackage{makecell}
\usepackage{multirow}
\usepackage{multicol}
\usepackage[lined,boxed,commentsnumbered, ruled,linesnumbered,resetcount]{algorithm2e}
\usepackage{algpseudocode}
\usepackage{multirow}
\usepackage{amsmath, amsthm, amssymb}
\usepackage{soul}
\usepackage{booktabs} 
\usepackage{amsfonts,amsmath,amssymb,amsthm}
\usepackage{verbatim,float,url,enumerate}
\usepackage{verbatim,float,url,enumerate}
\usepackage{subfigure,epsfig,psfrag}
\usepackage{bm,dsfont,color,appendix}
\usepackage{adjustbox}
\usepackage{mathtools}
\usepackage{bbm}
 \usepackage{booktabs}
\usepackage{amsmath}
\usepackage{latexsym}
\usepackage{setspace}
\usepackage{tikz}
\usetikzlibrary{arrows,positioning}
\usepackage{rotating}
\usepackage{hyperref}
\usepackage{makecell}
\usepackage{tabularx}
\usepackage{amsmath}  
\newtheorem{example}{Example}
\newcommand{\bP}{\mathbb{P}}
\newcommand{\bE}{\mathbb{E}}

\newcommand{\mI}{\mathcal{I}}
\newcommand{\mB}{\mathcal{B}}
\newcommand{\real}{\mathbbm{R}}
\newcommand{\mP}{\mathcal{P}}

\theoremstyle{plain}
\newtheorem{theorem}{Theorem}[section]
\newtheorem{proposition}[theorem]{Proposition}

\theoremstyle{definition}

\theoremstyle{remark}
\usepackage[normalem]{ulem}
\newtheorem{remark}[theorem]{Remark}

\usepackage[round]{natbib}

\begin{document}

%

%

\twocolumn[

\aistatstitle{Conformalized Semi-supervised Random Forest for Classification and Abnormality Detection}

\aistatsauthor{Yujin Han\textsuperscript{\textdagger} \And Mingwenchan Xu\textsuperscript{\textdagger} \And  Leying Guan}

\aistatsaddress{Department of Computer Science, \\The University of Hong Kong\\ Hong Kong, China \And  Department of IEMS\\ Northwestern University \\Illinois, USA\And Department of Biostatistics \\Yale University\\ New Haven, USA}]




\begin{abstract}
The Random Forests classifier, a widely utilized off-the-shelf classification tool, assumes training and test samples come from the same distribution as other standard classifiers. However, in safety-critical scenarios like medical diagnosis and network attack detection, discrepancies between the training and test sets, including the potential presence of novel outlier samples not appearing during training, can pose significant challenges. To address this problem, we introduce the Conformalized Semi-Supervised Random Forest (CSForest), which couples the conformalization technique Jackknife+aB with semi-supervised tree ensembles to construct a set-valued prediction $C(x)$. Instead of optimizing over the training distribution, CSForest employs unlabeled test samples to enhance accuracy and flag unseen outliers by generating an empty set. Theoretically, we establish CSForest to cover true labels for previously observed inlier classes under arbitrarily label-shift in the test data. We compare CSForest with state-of-the-art methods using synthetic examples and various real-world datasets, under different types of distribution changes in the test domain. Our results highlight CSForest's effective prediction of inliers and its ability to detect outlier samples unique to the test data. In addition, CSForest shows persistently good performance as the sizes of the training and test sets vary. Codes of CSForest are available at \href{https://github.com/yujinhan98/CSForest}{https://github.com/yujinhan98/CSForest}.
\end{abstract}
.\makeatletter\def\@makefnmark{}\makeatother
\footnotetext{ \textdagger Equal contribution. This work was done at Yale University. Correspondence to: leying.guan@yale.edu.}

\section{INTRODUCTION}
\label{Intro}

A classifier typically generates predictions for a test sample by choosing the class label associated with the highest predicted probability. This approach proves inadequate for addressing the increasing demand for assessing prediction reliability in practical scenarios, such as medical diagnosis \citep{esteva2017dermatologist, kompa2021second} and autonomous vehicles \citep{kalra2016driving, qayyum2020securing}. One approach for addressing this challenge involves minimizing a combined cost associated with misclassification and rejection, permitting the avoidance of predictions for test samples exhibiting high uncertainty. For example, if the maximum estimated probability $\max_{k\in \{0,1\}} \hat p(k|x)$ for the binary response is low where $\hat p(k|x)$ is the estimated probability of beging in class $k$ using the training data, we might choose not to predict a test observation $x$ \citep{chow1970optimum, herbei2006classification, bartlett2008classification}. This idea has been implemented across various learning algorithms and expanded to address multi-class classification problems \citep{cortes2016boosting,ni2019calibration,charoenphakdee2021classification}. The set-valued prediction via conformal prediction provides an alternative framework \citep{vovk2005algorithmic,papadopoulos2002inductive,lei2015distribution,gammerman2013learning}, in which the classifier generates a set covering all possible labels for a given observation $x$ based on the conformal score function $s(x,k)$ that measures the plausibility of label for $x$ being $k$, e.g., $s(x,k)\leftarrow \hat p(k|x)$. For instance, one can form the calibrated set-valued prediction set $C(x) = \{k: s(x,k) \geq \tau_k\}$, with $\tau_k$ being class-dependent and calculated to ensure a desired coverage of the true label \citep{vovk2005algorithmic}.

Traditionally, classification uncertainty quantification assumes that the training and test samples are independently and identically distributed (i.i.d.) from the same distribution. In reality, this assumption doesn't always hold. For instance, in medical applications, the test cohort may include samples representing novel pathologies that bear little similarity to the labeled training set \citep{lin2005approximations}. Similarly, network attackers may create new intrusions to evade existing detection systems \citep{marchette2001computer}. Therefore, it becomes essential to assess uncertainty under distributional changes and flag test samples where predictions should not rely solely on the model trained with the training data.

To address this challenge, we introduce CSForest (\textbf{C}onformalized \textbf{S}emi-Supervised Random \textbf{Forest}), an ensemble tree classifier that leverages both labeled training data and unlabeled test data to form calibrated set-valued prediction and flag outliers.  The term ``semi-supervision" here refers to the utilization of unlabeled test data. CSForest builds upon recent work on test-data optimized calibrated classification framework\citep{guan2019prediction}. \cite{guan2019prediction} constructs a calibrated semi-supervised set-valued prediction via sample-splitting where one subset of samples is used for training the model while the remaining part is for calibration. In contrast, CSForest avoids the sample-splitting schema, constructing the random forest tree ensembles and calibrating the prediction using all samples. We summarize our contributions into three main aspects:

\begin{enumerate}
     \item We present a novel classifier, CSForest, designed for classification with calibrated uncertainty quantification in the presence of distributional shifts between training and test datasets. It employs a novel semi-supervised random forest structure that differentiates between observed training classes and unlabeled test data, and adapts the conformalization technique Jackknife+aB \citep{kim2020predictive} to handle the case of joint and asymmetric utilization of both training and test samples. 
     \item We provide a theoretical guarantee for true lable coverage using $C(x)$ constructed by CSForest, under arbitrarily shifted test distributions. This theoretically ensures the effectiveness of CSForest under varying degrees of data drift.
    \item We conduct extensive experiments on simulated and publicly available datasets under various label shift settings to demonstrate CSForest's gain over existing state-of-the-art methods. 
\end{enumerate}

\section{RELATED WORK}
\label{sec:related}

\noindent\textbf{Distribution Shift.} 
 Regarding distributional changes, both the covariate and label shifts are commonly studied \citep{scholkopf2012causal}. The former assumes the conditional density $p(y|x)$ to be fixed with  $f(x)$, the marginal density of $x$, potentially changing \citep{shimodaira2000improving,bickel2009discriminative, gretton2009covariate, csurka2017domain}; the latter treats $f_k(x)$, conditional density of $x$ given the label $y=k$, as fixed, but the prevalence of different labels can vary among the observed training classes \citep{storkey2009training, lipton2018detecting}. 

Recently, \cite{guan2019prediction} proposes BCOPS, a test-data optimized calibrated classifier, and the Generalized Label Shift (GLS) model defined in eq.~(\ref{eq:glabelshift}), which extends the label shift model to include unseen classes to handle outliers. Suppose that the training data is a mixture of $K$ different classes. For class $k$, its mixture proportion is $\pi_k$, and feature density is $f_k(x)$, with $\pi_k$ satisfying $\sum_{k=1}^K \pi_k = 1$. The generalized label shift model assumes  a target distribution accepting both label shift among training classes and the appearance of outlier component(s) and requires only $f_k(x)$ to remain the same for each observed class:
  \begin{equation}
  \label{eq:glabelshift}
  \mu(x) = \sum_{k=1}^K \tilde\pi_k f_k(x)+\delta \cdot f_{R}(x),
  \end{equation}
  where $\delta+\sum_{k=1}^K \tilde \pi_k = 1$, $\tilde\pi_k\geq 0$ represents the proportion of samples from class $k$ in the target distribution, $\delta \geq 0$ represents the proportion of outlier samples not from the observed classes, and $f_R(x)$ represents the density for the outlier component. Under the GLS model, BCOPS utilizes both labeled training samples and unlabeled test samples to construct calibrated set-valued prediction. The crucial calibration step of BCOPS relies on sample-splitting, which results in low data-utilization efficiency, especially when training or test samples are limited.

\noindent\textbf{Conformal Prediction.} Conformal prediction (also known as conformal inference) \citep{vovk2005algorithmic, papadopoulos2002inductive, lei2015distribution} aims to create statistically rigorous uncertainty sets/intervals for the predictions from classical machine learning models, aiming to cover the true label with a desired probability in the non-asymptotic regime, without model assumptions on how $y$ depends on $x$. Consider  $(x_i,y_i)_{i=1}^n$ as $n$  (feature, label) pairs, and a new sample $(x_{n+1},y_{n+1})$ where $y_{n+1}$ is unobserved. Based on the previous $n$ observations, the conformal prediction creates a prediction set $\hat C_n(x_{n+1})$ for the new instance $x_{n+1}$ and ensure that $\mathbb{P}(y_{n+1} \in \hat C_n(x_{n+1})) \geq 1-\alpha$, where $\alpha \in (0,1)$ is the allowed miscoverage level. For example, in the classification setting, if $\alpha=0.1$, then the probability that $\hat C_n(x_{n+1})$ contains the true label $y_{n+1}$ is no smaller than 90\%. 

A key step in forming the conformal prediction is the choice of the conformal score function $s(x, y)$, which is used for evaluating how plausible of observing certain $(x_i, y_i)$, followed by a calibration of the probability for observing $(x_{n+1}, y_{n+1})$ via comparing its score to those from  labeled training samples. As some examples for classification problems, \citet{romano2020classification} considers $s(x, k)$ as the probability (estimated) of observing labels with estimated conditional probability $p(.|x)$ no worse than that of class $k$, and $\hat C_n(x_{n+1})=\{k: s(x_{n+1},k)\geq \tau\}$ where $\tau$ is a threshold determined by the empirical distribution $\{s(x_i,y_i)\}_{i=1}^n$. \cite{vovk2005algorithmic} and \cite{lei2014classification} consider the conformal score $s(x,k)$ as the conditional probability of having label $k$ or density of $x$  in class $k$, and construct $\hat C_n(x_{n+1}) = \{k: s(x_{n+1},k) \geq \tau_k\}$ using a class-specific threshold $\tau_k$ where $\tau_k$ is determined by the empirical distribution $\{s(x_i,k)\}_{i:y_i=k}$.

Different conformal prediction schemes have been developed in the literature. During the early times, split conformal prediction is usually adopted where we train the conformal score function $s(.)$ using one-fold of the data and perform calibration $\tau$ using the remaining \citep{vovk2005algorithmic}. In recent years, significant progress has been made in cross-conformal prediction to improve data utilization efficiency. \citet{vovk2018cross} proposed splitting data into multiple folds, calculating scores for each fold using score functions learned from the remaining data, and aggregating all scores for calibration. \citet{barber2021predictive} developed Jacknife+ for regression problems, which combines Jacknife with conformal prediction and constructs the prediction interval as
\[\hat C(x) = \left\{y:\frac{1}{n+1}\left(1+\sum_{i=1}^n\mathbbm{1}_{ \hat s^{i}(x,y)\geq \hat s^{i}(x_i,y_i)}\right)\geq \alpha\right\},
\]
where $\hat s^i(x, y)=|\hat m^{i}(x) - y|$ is the conformal score function using the mean-prediction function $\hat m^i(x)$ learned from training samples excluding $(x_i, y_i)$. Although Jacknife+ can only provide a worst-case coverage guarantee at level $(1-2\alpha)$, the achieved empirical coverage is often well-calibrated. \citet{kim2020predictive} described Jacknife+aB to mediate the computational burden of Jacknife+, which ensembles and calibrates prediction using repeated Bootstraps rather than retraining the model after excluding each training sample.

\begin{figure*}
    \centering
   \includegraphics[height = .25\textwidth, width=0.78\textwidth]{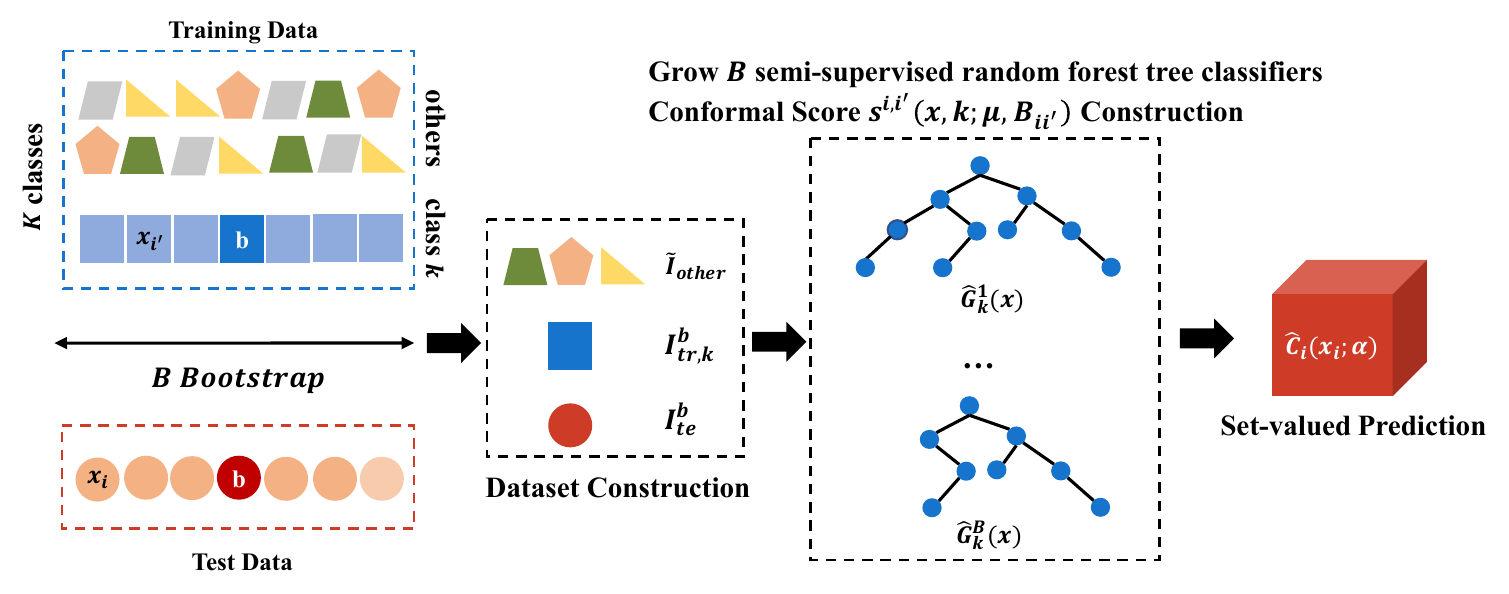}
    \caption{Overview of CSForest. For class $k$, let $\mI^b_{k}$, $\mI^b_{te}$ and $\tilde\mI_{other}$ be Bootstrap samples from   from training class $k$, test samples and  training samples other than class $k$. We train a multi-class tree classifier with random feature selection as in the random forest using the Bootstrapped samples,  where we maintain the labels all training samples and treat the test set as its own class. The resulting $B$ random forest tree classifiers, $\{\hat G^1(x),..., \hat G^B(x)\}$,  are used to separate different labeled classes and the test samples. For the sample pair $x_i\in \mI_{te}$ and $x_i'\in \mI_{k}$, we aggregate trees that do not use $x_i$ and $x_i'$ (i.e., the data $\mB_{ii'}=\{b: i\notin \mI_{te}^{b}, i'\notin \mI^{b}_k\}$) to form an ensemble classifier, and subsequently, an ensemble conformal score function $\hat s^{ii'}(x, k;\mu)$ . Finally, we use the score function $\hat s^{ii'}(x, k; \mu)$ and compare $\hat s^{ii'}(x_i, k; \mu)$ to $\hat s^{ii'}(x_{i'}, k; \mu)$ for all $i' \in \mathcal{I}_{k}$ to form the calibrated evaluation $\hat s_{ik}$ for test sample $x_i$ being in class $k$ and include $k$ in the prediction set $\hat{C}(x_i)$ if $\hat s_{ik}$ is no smaller than $\alpha$.
    }
    \label{fig:StructureCSForest}
\end{figure*}
   
\section{CONFORMALIZED SEMI-SUPERVISED RANDOM FOREST}
\label{method}
Despite the popularity of random forest and its variants, existing work implicitly assumes that training and test samples originate from the same distribution. This reliance makes them unreliable in the presence of distributional changes, which can be particularly problematic in safe-critical applications. Classification uncertainty quantification in this setting is also challenging. To address this issue, we introduce CSForest (\textbf{C}onformalized \textbf{S}emi-Supervised Random \textbf{Forest}), a tree-ensemble classifier that produces set-valued predictions designed to incorporate true labels while minimizing the inclusion of false labels, customized to match a target distribution $\mu(x)$.
\begin{align}
    &\min \int_{x} |C(x)|\mu(x) d x,   \label{eq:GLS}\\
    &s.t.\; \bP[k\in C(X)|Y=k] \geq 1-\alpha,
    \mbox{for all }k=1,\ldots, K.\notag
\end{align}
Specifically, CSForest optimizes for a target distribution as a mixture of the training
density $f_{tr}(x)$ and test feature
density $f_{te}(x)$ and set $\mu(x) = f_{te}(x)+ wf_{tr}(x)$, where $w \geq 0$. If $w = 0$, $\mu(x) = f_{te}(x)$ and the objective of CSForest coincides with the objective of BCOPS, which optimizes for the expected test cohort classification accuracy. On the other hand, when $w$ is large, it has a similar objective as the CRF model and optimizes classification performance on samples generated the same way as the training cohort. When $w$ is not excessively large, $f_{te}(x)$ is a significant contributor to $\mu(x)$, the constrained optimization objective in eq.~(\ref{eq:GLS}) encourages $C(x) = \emptyset$ for unseen outliers even though we do not explicitly model outliers. In other words, if $x$ is unlikely to belong to any of the observed training classes, we prefer to classify it as an outlier.

The following Proposition \ref{prop:oracle} further provides the oracle solution to eq.~(\ref{eq:GLS}) under GLS model:
\begin{proposition}
\label{prop:oracle}
Set the conformal score function as $s(x, k;\mu) = [f_k(x)\slash\mu(x)]$. Under the GLS model, the solution to eq.~(\ref{eq:GLS}) is $C(x) =\{k: \bE_X[\mathbbm{1}\{s(x,k; \mu)\geq s(X,k;\mu)\}|Y=k]\geq \alpha,k=1,\ldots, K\}$.
\end{proposition}
Proposition \ref{prop:oracle} rephrases Proposition 2 from \cite{guan2019prediction}, and its proof is provided in Appendix \ref{Proof of Theorem 1} for completeness. CSForest estimates $C(x)$ in Proposition \ref{prop:oracle} using a semi-supervised random forest that utilizes labeled training samples and the unlabeled test cohort, coupled with the Jackknife+aB strategy for correct calibration to ensure coverage guarantee. Specifically, for a test sample $x_{i}$ and a training sample $x_{i'}$ from class $k$ with size $n_k$, CSForest constructs a conformal score function $\hat s^{ii'}(x,k;\mu)$  trained without $x_{i}$ and $(x_{i'},y_{i'})$ to measure how likely a sample is from class $k$. More specifically, CSForest replaces the oracle score $s(x,k;\mu)$ in Proposition \ref{prop:oracle} when comparing $x_{i}$ and $x_{i'}$ and replaces expectation $\bE_X[.]$ is replaced by an empirical version using corresponding  $\hat s^{ii'}(x,k;\mu)$. In other words, let $n_k$ be the training sample size for class $k$, the inclusion criterion in  Proposition \ref{prop:oracle} is replaced by its empirical version below:
\begin{equation}
\label{eq:CSForest_score}
     \hat{s}_{ik}=\frac{1+\sum_{y_i=k}\mathbbm{1}\{\hat s^{ii'}(x_{i},k;\mu)\geq \hat s^{ii'}(x_{i'},k;\mu)\}}{n_k+1}.
\end{equation}
The the estimated prediction set $\hat C(x_i)$ is :
\begin{equation}
\hat C(x_i) = \{k: \hat s_{ik}\geq \alpha\}.
\end{equation}

\begin{figure*}
        \centering
         \includegraphics[height = .25\textwidth]{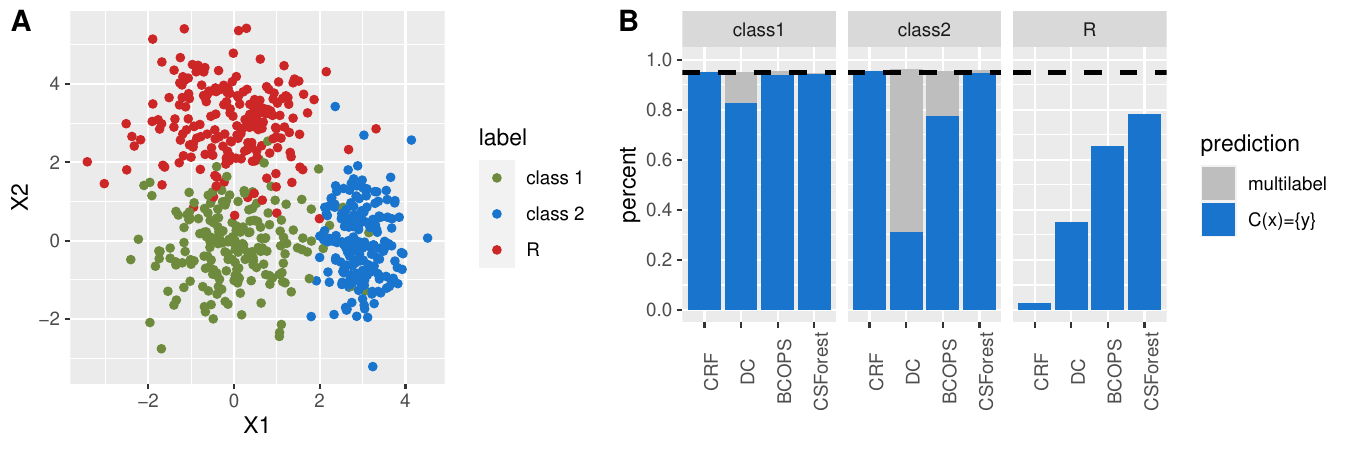}
        \caption{Panel A shows the first two dimensions of samples are generated from the three classes: green\slash blue\slash red points representing samples from class 1\slash 2\slash R.  Panel B shows the coverage rate which is defined by the proportion of samples with true labels included in their prediction sets. The horizontal dash line refers to the coverage level of 95\%. Panel B is grouped by the actual labels in the testing data and colored based on if a prediction set contains only the correct label (blue) or more than the correct label (gray). }
          \label{fig:illustration}
    \end{figure*}

Given a user-specified weight $w$, Figure \ref{fig:illustration} presents a graphical illustration of its model structure and  Algorithm \ref{alg:CSForestI} delineates the ensemble tree constructions and prediction calibrations for CSForest. In Algorithm \ref{alg:CSForestI}, we use $\mI_{tr}$ and $\mI_{te}$ to denote the training and test sets, respectively, and $\mI_{k}$ to denote samples from the training class $k$. For each class $k$, lines 2-5 construct $B$ Bootstrapped random forest tree classifiers  to separate the training classes and the test samples.  The random forest tree refers to a tree whose split is selected by the best split from $L$ randomly selected candidate features, as constructed in the random forest, with $L=\lfloor\sqrt{p}\rfloor$ as the default split number in the R {\it range} package. Line 6 constructs the conformal score function $\hat s^{ii'}(x,k;\mu)$, using only the trees excluding test sample $x_i$ and the training sample $x_{i'}$. {In short, Algorithm \ref{alg:CSForestI} can be seen as estimating the oracle conformal score $s(x, k;\mu)$ under the target distribution with $w$ being not excessively large, by utilizing trees from a weighted random forest classifier.}

The estimated conformal score functions are then used to construct the calibrated score, $\hat s_{ik}$, and the prediction set, $\hat C(x)$, in lines 8-13. It is worth noting that, in line 3, to prevent redundant resampling in Bootstrap, we constrain $\tilde\mI_{other}$ to be the Bootstrap sample of size $\min(\lceil n_{te}w \rceil,n-n_k)$ drawn from training samples, excluding class $k$. It is worth noting that the probability $Pr(B_{ii'}=\emptyset)$ decreases rapidly as $B$ increases.

\begin{algorithm}
\caption{CSForest}
   \label{alg:CSForestI}
\SetKwInOut{Input}{Input}
\SetKwInOut{Output}{Output}
\setcounter{AlgoLine}{0}
\Input{Training Data $\{(x_i, y_i),i\in \mI_{tr}\}$,  Test Data $\{x_i,i\in \mI_{te}\}$, $\tilde B$ and $w$ (1 by default.)}
\Output{Prediction sets $\hat C_i(x_i)$ for $i\in \mI_{te}$.}
\For{$k=1,\ldots,K$}{
Sample $B$ from ${\rm{Binomial}}(\tilde B; (1-\frac{1}{n_k+1})^{n_k})$.
\For{$b=1,\ldots, B$}{

Let $\mI_{k}^{b}$, $\mI_{te}^{b}$ be the Bootstraps of $\mI_{k}$ (index of training class $k$) and $\mI_{te}$. Let $\tilde\mI_{other}$ be the Bootstrap of size $\min(\lceil n_{te}w \rceil,n-n_k)$ from the remaining training sample indices $\mI\setminus \mI_k$.

Grow a single random forest tree classifier $\hat G^b(x)$ separating different labeled classes and the test samples using $\mI_{k}^{b}\cup\mI_{te}^{b}\cup \tilde\mI_{other}$.}

For sample pair $i\in \mI_{te}, i'\in \mI_{k}$, set $\mB_{ii'}=\{b: i\notin \mI_{te}^{b}, i'\notin \mI^{b}_k\}$ and construct the conformal score function $\hat s^{ii'}(x,k;\mu)=\left(\sum_{b\in \mB_{ii'}}\hat G^b_k(x)\right)\slash |\mB_{ii'}|$.}

\For{$i\in \mI_{te}$}{
\For{$k=1,\ldots, K$}{

Construct $\hat s_{ik}$ for sample $i$ and class $k$ via eq.~(\ref{eq:CSForest_score}).

}
Construct $\hat C(x_i) = \left\{k: \hat s_{ik}\geq \alpha\right\}$.

}
\end{algorithm}

\begin{remark}
\label{remark:emptyB}

When $B_{ii'}$ is empty, $s^{ii'}(x,k;\mu)$ is not defined. In this case, We will exclude training sample $x_{i'}\in \mI_k$ when calibrating for test sample $x_i$. Let $n_{te}$ be the size of $\mI_{te}$. The probability that $B_{ii'}=\emptyset$ is bounded by $\bP[i'\in \mI_{k}^b, \forall b\leq B]+\bP[i\in \mI_{te}^b, \forall b\leq B]=[1-(1-\frac{1}{n_{k}})^{n_k}]^B+[1-(1-\frac{1}{n_{te}})^{n_{te}}]^B\approx 2(1-\frac{1}{e})^B$ for decently large $n_{te}$ and $n_k$, which decreases fast with $B$.
\end{remark}

Theorem \ref{thm:coverage} states that CSForest provides a worst-case coverage guarantee for the true response at the level $(1-2\alpha)$.
\begin{theorem}
\label{thm:coverage}
Suppose the generalized label shift model holds where features from class $k$ are i.i.d generated from a distribution $\mP_k$. For any fixed integers $\tilde B \geq 1$,  the constructed $\hat C_i(x)$ from CSForest satisfies:
\begin{align}
\label{eq:coverage}
&\bP\left[k\in \hat C(x_i)|y_i=k\right]\geq 1-2\alpha, \\
&\mbox{for all } i\in \mI_{te}\; \mbox{ and }\; k = 1,\ldots, K. \nonumber
\end{align}
\end{theorem}
\vskip -.1in

While the proof relies on the previous arguments used in Jacknife+aB for supervised regression problem \cite{kim2020predictive}. However, new conditioning arguments  are needed to estabilish exchangeability due to the paired sampling of both training and test samples. Please find the proof of Theorem \ref{thm:coverage} in Appendix \ref{Proof of Theorem 1}. 

Theorem \ref{thm:coverage} ensures per-class coverage for observed training classes, which means we guarantee true label coverage for inlier classes even in arbitrarily shifted test distributions, e.g.,
 \[
    \bP\left[y\in C(x)| y\in \{1,\ldots, K\}, (x, y)\sim P_{te}\right] \geq 1-2\alpha, 
\]
for any test distribution $P_{te}$ satisfying the generalized label shift model assumption. Although the theoretical guarantee for the worst-case coverage is at the level $(1-2\alpha)$, the empirical coverage using CSForest is usually close to or above the targeted level $(1-\alpha)$.

\begin{figure*}
  \centering
  \renewcommand\thesubfigure{(\Alph{subfigure})}
  \subfigure[Per-class quality evaluation with outliers but no additional label shift among inlier digits, where the outliers are defined as $R = \{6,7,8,9\}$. ]{
\includegraphics[height=0.25\textheight,width=0.49\textwidth]{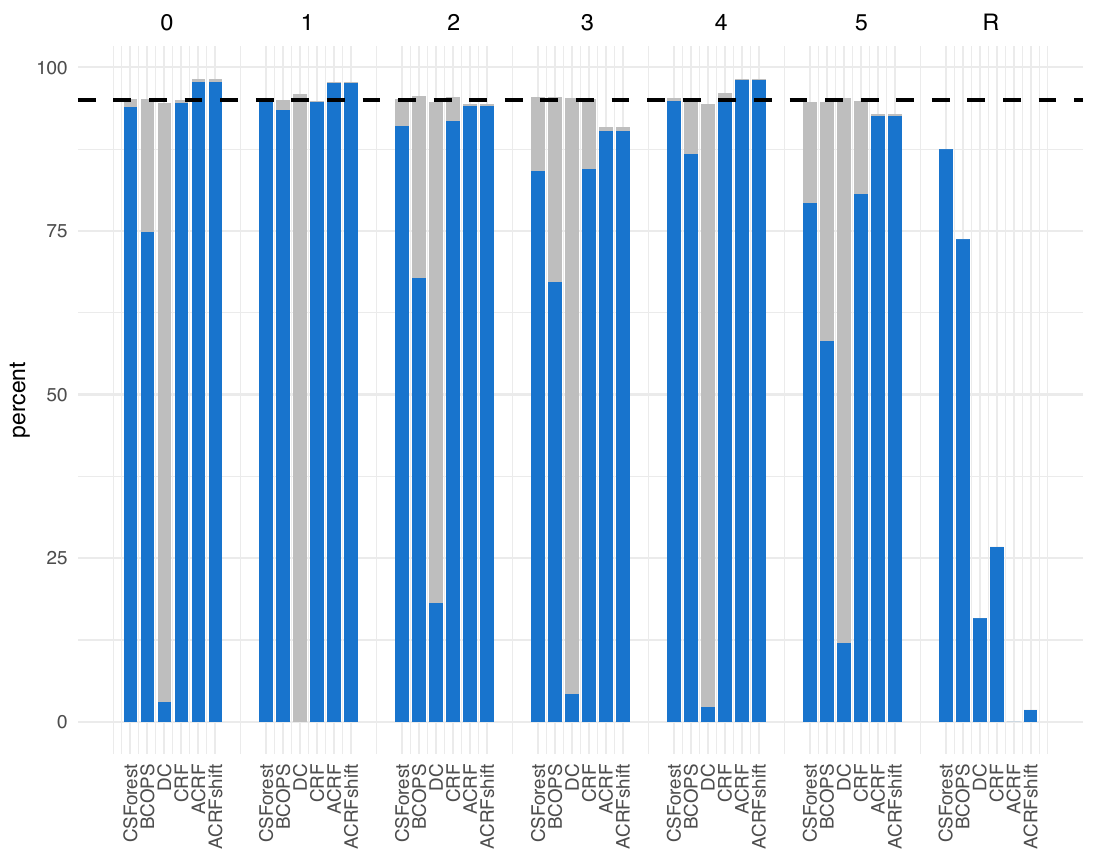}
    \label{fig:CoverageRateBarMNISTI}
  }\quad
  \subfigure[Per-class (class 0-5) quality evaluation with additional label shift among inlier digits but no outliers.]{\includegraphics[height=0.25\textheight,width=0.45\textwidth]{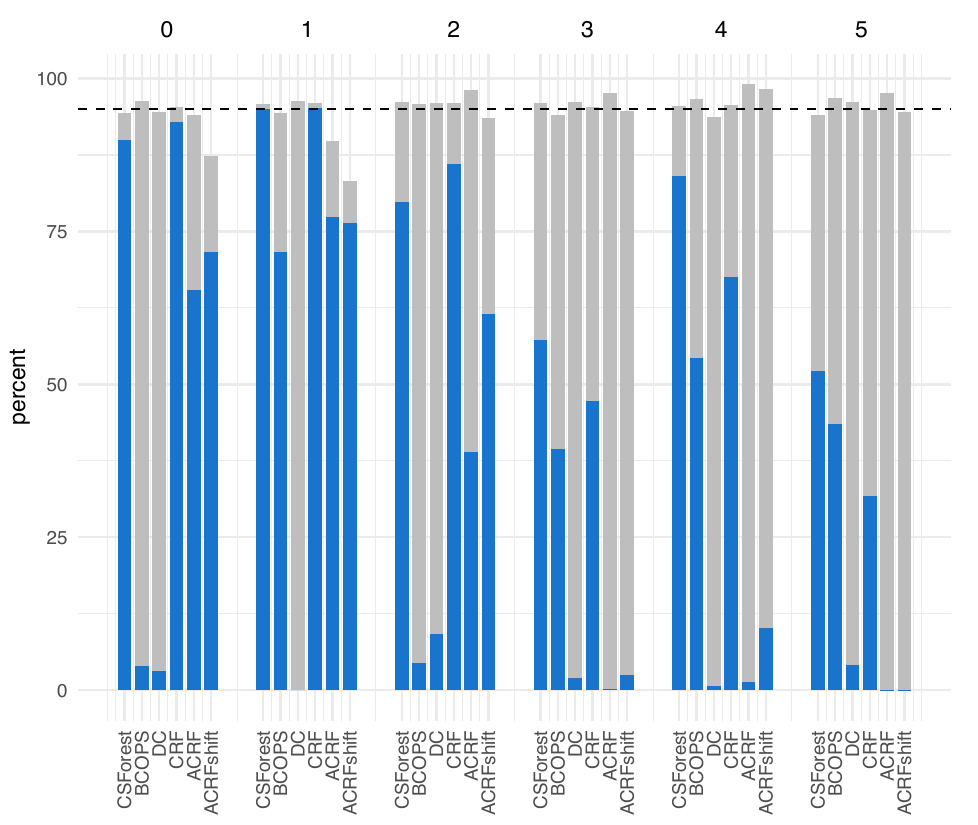}
    \label{fig:CoverageRateBarMNISTI_No}
  }
  \vskip -.1in
  \caption{Per-class quality evaluation on MNIST. Panel A and B were grouped by the true labels in the testing data and colored based on whether a prediction set contains only the correct label (blue) or more than the correct label (gray). The horizontal dash line refers to the coverage level of 95\%.}
  \label{fig:per_class_mnist}
\end{figure*}

\section{EXPERIMENTS}
\label{sec:experiments}

\subsection{Synthetic Data}
\label{Synthetic Data}

We begin with a simple illustrative 2D synthetic dataset and compare the performance using CSForest and three closely related set-valued conformal classifiers:  BCOPS, which is a conformalized semi-supervised classifier that uses half of the samples for training while the other half for calibration (see Section \ref{sec:related}); DC (density set classifier)\citep{cadre2006kernel,lei2014classification,hechtlinger2018cautious,sadinle2019least} and CRF (conformalized random forest), which follow the sample-splitting conformal prediction scheme while using the kernel estimate of the per-class density for $x|y$ and the estimated conditional probability of observing a label  $y|x$ via a random forest on the training samples as the conformal score functions, respectively.
\begin{example}
 \label{example:intro}
 Let $X\in \real^{10}$ be the feature. We observe two classes $Y\in \{1,2\}$ in the training data, but the test data contains outliers labeled with $Y=R$. We generate $X_j\sim N(0,1)$ $(j=3,\ldots, 10)$ as noise, with different classes separated by the first two dimensions:
 {\small
 $$\left\{
    \begin{aligned}
    & X_1 \sim N(0,1), \;X_2\sim N(0,1), \quad Y = 1,\\
    &X_1\sim N(3,0.5), \;X_2\sim N(0,1),  \quad  Y = 2,\\
    &X_1\sim N(0,1), \;X_2\sim N(3,1),  \quad Y = R.
    \end{aligned}
    \right.
    $$
    }
 \end{example}
\vskip -.13 in
Figure \ref{fig:illustration}(A) shows the first two dimensions of samples generated from the three classes $Y\in \{1,2, R\}$.  We generated 200 samples from classes 1 and 2 to form the training set and 200 samples from each of the three classes to form the test set. In Figure \ref{fig:illustration}(B), we evaluated the quality of the set-valued prediction $\hat{C}(x)$  using DC, CRF, BCOPS, and CSForest across 20 independent runs with a targeted miscoverage rate at $\alpha =0.05$. All four methods achieve the desired 95\% $(1-\alpha)$ coverage on true labels. However, both CSForest and BCOPS adapted to the test cohort and outperformed CRF and DC significantly in outlier detection. Additionally, compared to BCOPS, CSForest had fewer samples with multiple labels from classes 1 and 2 and exhibited a higher rejection rate for outliers.

\subsection{Real-World Data}
In this section, our primary objective is to evaluate the effectiveness of CSForest on various real-world datasets, focusing on addressing the following three questions:

\begin{enumerate}
    \item[Q1.] Can CSForest detect outliers efficiently while making accurate predictions for inliers in the presence of outliers but no additional label shift among inliers? (denoted as \textbf{outliers w/o shift}.)
    
    \item[Q2.]  Under the traditional label shift model without outliers, can CSForest achieve competitive performance compared to alternative classifiers? (denoted as \textbf{shift w/o outliers}.)
    
    \item[Q3.] Does CSForest demonstrate stable performance as the training and test sample sizes vary?
\end{enumerate}

Q1 and Q2 capture two extreme settings of the GLS model. We have set $w$ at its default value for all experiments in the main paper with $w=1$. In Appendix \ref{Discussion on w}, {we include the sensitivity analysis of the crucial parameter $w$ and shows CSForest performs well for different values of $w$, with $w = 1$ being a reasonable choice to balance performance for both inliers and outliers}.

\begin{table*}
\caption{Achieved type I and type II errors at $\alpha = 0.05$ under different distributional shift settings. While most methods achieved desirable type I errors and true label coverage rates ($1-\alpha$), only CSForest consistently achieved lower type II errors in both settings.}
\centering
\begin{tabular}{lccccccc}
\toprule
\multirow{2}{*}{Dataset} & \multirow{2}{*}{Method} & \multicolumn{2}{c}{outliers w/o shift}        & \multicolumn{2}{c}{shift w/o outliers}                \\ \cmidrule(lr){3-4} \cmidrule(lr){5-6}  & & Type I Error&Type II Error &Type I Error&Type II Error\\ 
\midrule
\multirow{6}*{MNIST}&CSForest&   0.049$\pm$0.006   &    \textbf{0.091 $\pm$ 0.008}&  0.048$\pm$0.016 & \textbf{0.291$\pm$0.038}   \\ 
&BCOPS&0.048$\pm$0.004   &    0.237$\pm$0.019&     0.042$\pm$0.007   &    0.556$\pm$0.040  \\ 
&DC  &     0.049$\pm$0.008  &    0.890$\pm$0.021&     0.046$\pm$0.016   &    0.968$\pm$0.022   \\ 
&CRF   &     0.048$\pm$0.007  &    0.338$\pm$0.035  &     0.046$\pm$0.018   &    0.428$\pm$0.082  \\ 
&ACRF&   {0.046$\pm$0.006}   &   0.430$\pm$0.003 &    {0.025$\pm$0.011}    &   0.884$\pm$0.012\\
&ACRFshift &   {0.046$\pm$0.006}    &   0.432$\pm$0.009&    0.055$\pm$0.013    &  0.828$\pm$0.015 \\
\midrule
\multirow{6}*{CIFAR-10}&CSForest&   0.051$\pm$0.008   &    \textbf{0.000$\pm$0.000} &     0.049$\pm$0.013 &    0.009$\pm$0.035  \\ 
&BCOPS&0.049$\pm$0.006   &    0.001$\pm$0.000&     0.042$\pm$0.009   &    0.029$\pm$0.006     \\ 
&DC  &     0.046$\pm$0.007  &    0.048$\pm$0.091 &     {0.039$\pm$0.010}   &    0.071$\pm$0.115   \\ 
&CRF   &     0.049$\pm$0.008  &    0.003$\pm$0.000  &     0.047$\pm$0.015   &    \textbf{0.000$\pm$0.000}      \\ 
&ACRF&{0.003$\pm$0.001}   &   0.402$\pm$0.001&    0.040$\pm$0.009    &   0.221$\pm$0.023 \\
&ACRFshift & {0.003$\pm$0.001}    &   0.069$\pm$0.003&    0.046$\pm$0.007    &  0.230$\pm$0.035 \\
\midrule
\multirow{6}*{FashionMNIST}&CSForest& 0.050$\pm$0.005   &    \textbf{0.266$\pm$0.018} &  {0.038$\pm$0.009} &    \textbf{0.311$\pm$0.040}    \\ 
&BCOPS&0.050$\pm$0.007   &    0.381$\pm$0.020   &     {0.038$\pm$0.009} &    \textbf{0.311$\pm$0.040} \\ 
&DC  &     0.051$\pm$0.007  &    0.666$\pm$0.033&     0.038$\pm$0.013   &    0.584$\pm$0.066    \\ 
&CRF   &     0.051$\pm$0.006  &    0.514$\pm$0.021 &     0.038$\pm$0.014   &    0.804$\pm$0.080    \\ 
&ACRF&    0.051$\pm$0.006   &   0.537$\pm$0.013&    0.054$\pm$0.009    &   0.835$\pm$0.020 \\
&ACRFshift &{0.046$\pm$0.005}  &   0.481$\pm$0.019 &    0.072$\pm$0.021    &  0.814$\pm$0.039  \\
\bottomrule
\end{tabular}
\label{tab:all results}
\end{table*}

\textbf{Datasets and Baselines.} Our evaluation is conducted on three well-established image benchmarks: MNIST \citep{lecun-mnisthandwrittendigit-2010}, FashionMNIST \citep{xiao2017fashion}, and CIFAR-10 \citep{krizhevsky2009learning}. {Additionally, we have included tabular data from the Network Intrusion domain and Chest X-ray data from the medical domain in the Appendix \ref{more data} to demonstrate the effectiveness of CSForest in handling diverse datasets. } To evaluate CSForest's performance, we compared it with BCOPS, CRF, DC along with two other approaches based on adaptive classification \citep{romano2020classification} and the covariate shift conformal prediction \cite{tibshirani2019conformal}: ACRF (Adaptive classifier via random forest) and ACRFshift (Adaptive classifier via random forest under covariate shift). ACRF is a derandomized version of the existing conformalized adaptive random forest classifier \citep{romano2020classification}, denoted as ACRFrandom, which aims for adaptive coverage across different feature regions. We will show results using ACRF instead of ACRFrandom in the main paper due to the latter's tendency to produce overly wide prediction sets, due to the attempt to achieve conditional coverage as indicated in the original paper \citep{romano2020classification}. ACRFshift combines ACRF with the covariate shift conformal prediction.  BCOPS, ACRF and ACRFshift all utilize random forest  for constructing set-valued predictions, as a fair comparison to CSForest. More details  of these baselines are provided in Appendix \ref{More details on baselines}.

\textbf{Training Details and Evaluations.}   We set the number of trees $B = 3000$ for CSForest, and repeat all experiments ten times for performance evaluations. We evaluate the effectiveness of all methods using the type I error, type II error, and the average set length of $\hat C(x)$ at $\alpha = 0.05$. Type I error is the percentage of samples with the true label excluded from their associated set-valued prediction $\hat C(x)$ for observed classes. This error measure is directly linked to the coverage guarantee in Theorem \ref{thm:coverage}. Type II error is calculated as the percentage of samples with $\hat C(x)$ containing labels other than the true labels. The average length $\hat C(x)$ under the mixed distribution $\mu(x)$ is the optimization objective under the formulation described by eq.~(\ref{eq:GLS}).

\subsubsection{The Outliers w/o Shift Setting}

 In this section, the test set contains outlier labels relative to the training set but without any additional label shift. For each data set, we constructed the training set by including a subset of class labels and the test set with all labels, e.g., for the MINIST data, the training set had digit labels 0-5 of equal size and the test set included both digit labels 0-5 of equal size and digit labels 6-9. Details of the data subsampling schemes for Q1 can be found in Appendix \ref{Training Details}.

Table \ref{tab:all results} displays the average type I and type II errors of all methods on the test data at $\alpha=0.05$. All methods achieved the the targeted coverage rate at $95\%$ when averaging inlier data.  Figure \ref{fig:CoverageRateBarMNISTI} presents detailed classification results with different methods on the MNIST dataset. Although ACRF, ACRFshift, and CRF achieved slightly higher accuracy compared to CSForest among inlier digits under no additional label shift, CSForest demonstrated the strongest capability to detect outlier digits 6-9, even compared to BCOPS, with an outlier detection accuracy of approximately 90\%. Similar results are observed on other datasets, as detailed in Appendix \ref{Per-class quality evaluation}.

\begin{figure*}
  \centering
  \includegraphics[height=0.35\textheight]{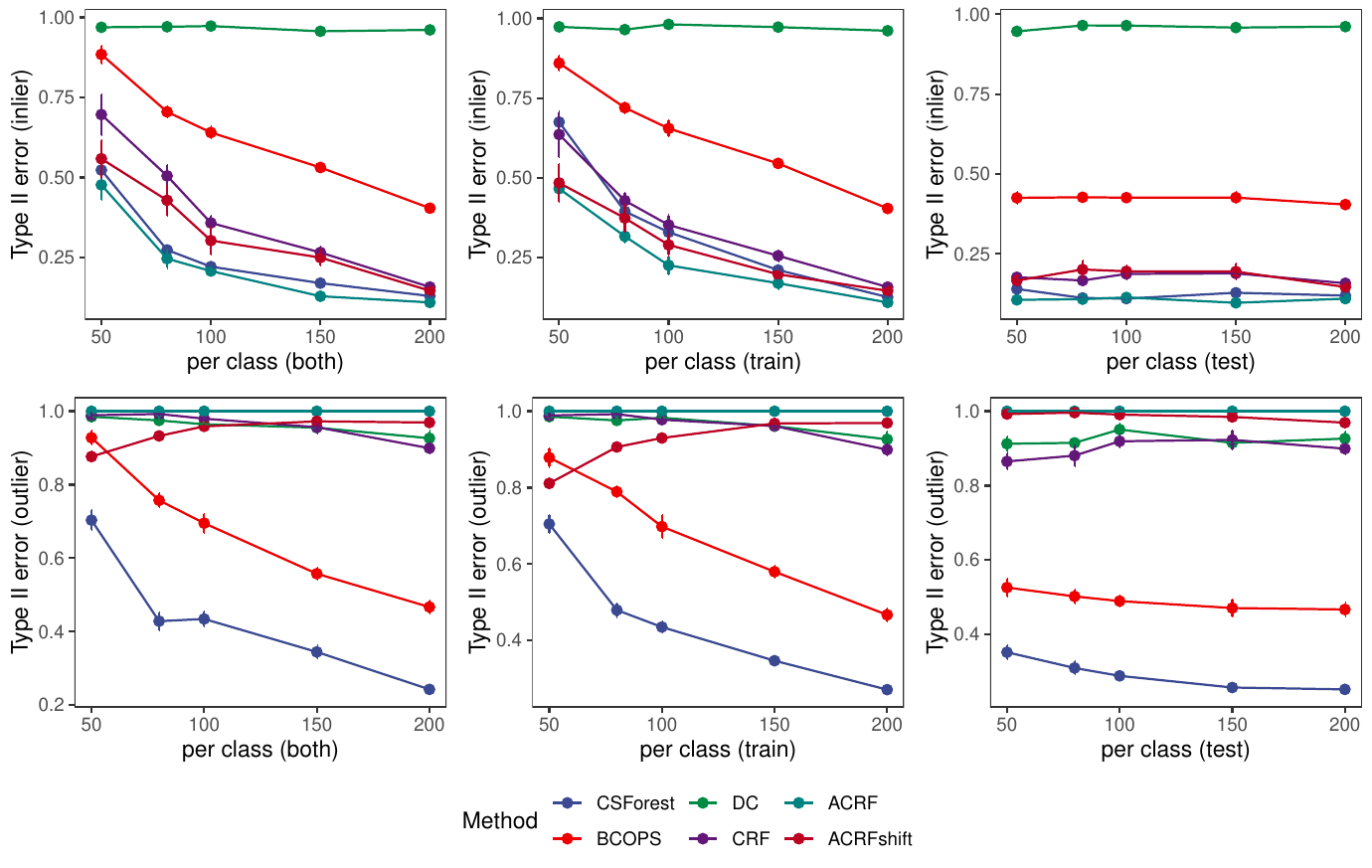}
  \caption{The type II error for inliers and outliers across different sample sizes on MNIST. Figure \ref{fig:MNISTvaringsize} demonstrates that CSForest outperforms the baselines by efficiently detecting outliers while maintaining lower inlier type II errors across various sample sizes. Note that error bars here are calculated based on repeated sample-splitting and can be smaller than the standard deviation due to sample dependence from different runs.}
  \label{fig:MNISTvaringsize}
\end{figure*}

\subsubsection{The Shift w/o Outliers Setting}
\label{sec:labelshift}

In the previous simulations, although we had outliers, the class ratios among inlier classes were balanced and remained the same for training and test data. To verify whether methods like CSForest, designed to achieve per-class coverage rather than marginal coverage, still maintain robustness when handling label shifts among inlier classes, we examine the predictive performance of all methods under the traditional label shift setting in the absence of outliers. For example, for the MNIST data set, the training and test sets contain digit labels 0-5 but with different class proportions. Details of the data subsampling and label shift schemes for Q2 can be found in Appendix \ref{Training Details}.

The achieved type I and type II errors using different methods in this standard label shift simulation can be found in Table \ref{tab:all results}. We observed that all methods achieved the desired coverage ($1-\alpha$) with ACRFshift exhibiting high variability. However, CSForest is the only method that achieved consistently low type II errors: CRF has a type II error 10\% and 50\% more than CSForest on MNIST and FashionMNIST, respectively; BCOPS has a type II error 25\% more than CSForest on MNIST; ACRF and ACRFshift both have type II errors more than two folds than those from CSForest on all three data sets. Figure \ref{fig:CoverageRateBarMNISTI_No} presents a detailed view of the prediction results for each class on the MNIST dataset. CSForest contains a higher proportion of samples with only the correct labels in almost every class compared to the baseline models, which underscores the high-quality prediction sets $\hat{C}(x)$ generated by CSForest under Q2, consistent with the lower type II error reported in Table \ref{tab:all results}. Detailed results on other datasets can be found in Appendix \ref{Per-class quality evaluation}.

\subsubsection{Comparisons with Varying Sample Sizes}

We conducted a comparison of different methods under varying sample size settings. Specifically, we varied the number of training and test samples per class from 50 to 200. Figure \ref{fig:MNISTvaringsize} presents the type II errors for inliers and outliers across all models on the MNIST dataset. In Figure \ref{fig:MNISTvaringsize}, it is evident that as training sample size increases, the type II error (inliers) decreases for all methods, while BCOPS and CSForest also benefit from increased test sample sizes. CSForest closely matches the CRF, the best-performing classifier in the inlier classification, for predicting inlier labels as we vary the training/test sample sizes from 50 to 200. CSForest and BCOPs outperformed other methods by a large margin for varying sample sizes for outlier detection, with CSForest significantly improving over BCOPS due to the enhanced data utilization efficiency. Of note, the ability for outlier detection (higher type II error for outliers) ACRFshift deteriorated as sample size increased. This surprising phenomenon is attributed to  ACRFshift's decision rule for outliers, which strongly depends on the sample weights under the covariate shift model, denoted as $\gamma_{x_0}(x) = \frac{r(x)}{r(x_0)+\sum_{z_i\in \mI_{cal}} r(x_i)}$. A sample is claimed an outlier $\gamma_{x_0}(x)$ is very large, and $\gamma_{x_0}(x)$ tends to increase with increased training sample sizes (Appendix \ref{More details on baselines}). Results on other datasets are consistent with those on MNIST (Appendix \ref{Per-class quality evaluation}).

\section{DISCUSSION}
\label{Discussion}
We propose CSForest, which aims to construct a calibrated and narrow set-valued prediction set under distributional changes, as a powerful ensemble classifier for robust inlier classification and outlier detection. We theoretically justified its robustness for covering the true class label and confirmed its ability to construct high-quality prediction sets compared to alternative methods via extensive experiments.

\begin{figure}[H]
    \centering
   \includegraphics[width=0.45\textwidth]{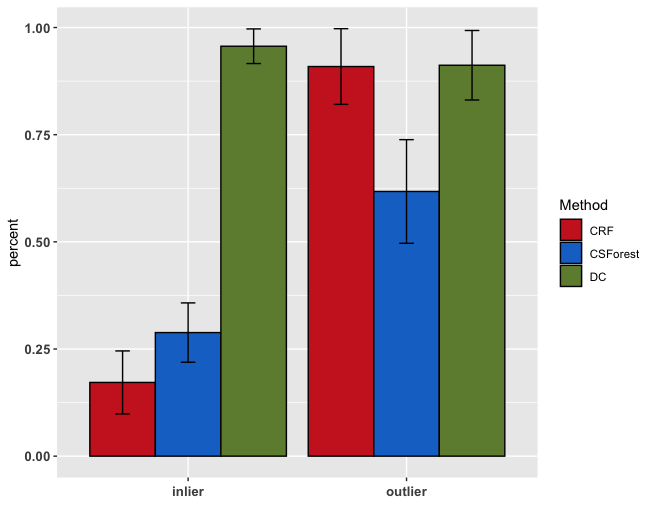}
   \caption{Achieved Type II errors for inliers and outliers across 100 repetitions at $\alpha = 0.05$ with merely 5 samples per-class in the test cohort.}
    \label{fig:MNISTerror}
  \end{figure}

\textbf{Future Work.} An interesting question is how much guidance from test samples is needed for effective outlier detection. Can CSForest still utilize the limited test samples for efficient outlier detection? As an exploratory experiment, we consider a challenging MNIST example, where we have 200 samples per class for labels 0-5 in the training set but only five samples per class for labels 0-9 in the test set.  Figure \ref{fig:MNISTerror} shows CSForest achieves an average type II error of approximately 42\% for inliers and 60\% for outliers, whereas DC exhibits an average type II error of as high as 95\%. This highlights the benefits of utilizing a small test set in CSForest and suggesting the potential of extending the framework of CSForest to settings with extremely limited or even single test samples for future exploration. 

Additionally, the GLS model assumes that the distribution of $x|y$ remains unchanged, which could be violated in practice.  When both the distribution of \(y\) and the distribution of \(x|y\)  are allowed to change, the problem becomes much more challenging and less well-defined. One interesting future direction is to relax GLS model and assume bounded small changes in \(x|y\), leading to the investigation of CSForest under an adversarial setting that allows adversarial yet small perturbations in \(x|y\) during test time.

\subsubsection*{Acknowledgements}

This work was supported by NSF award DMS2310836. 
\newpage
\bibliography{aistats2024}

\clearpage 
\newpage
\section*{Checklist}



 \begin{enumerate}

 \item For all models and algorithms presented, check if you include:
 \begin{enumerate}
   \item A clear description of the mathematical setting, assumptions, algorithm, and/or model. [Yes/No/Not Applicable] \textcolor{blue}{Yes}
   \item An analysis of the properties and complexity (time, space, sample size) of any algorithm. [Yes/No/Not Applicable] \textcolor{blue}{Yes}
   \item (Optional) Anonymized source code, with specification of all dependencies, including external libraries. [Yes/No/Not Applicable] \textcolor{blue}{Not Applicable}
 \end{enumerate}

 \item For any theoretical claim, check if you include:
 \begin{enumerate}
   \item Statements of the full set of assumptions of all theoretical results. [Yes/No/Not Applicable] \textcolor{blue}{Yes}
   \item Complete proofs of all theoretical results. [Yes/No/Not Applicable] \textcolor{blue}{Yes}
   \item Clear explanations of any assumptions. [Yes/No/Not Applicable]  \textcolor{blue}{Yes}
 \end{enumerate}

 \item For all figures and tables that present empirical results, check if you include:
 \begin{enumerate}
   \item The code, data, and instructions needed to reproduce the main experimental results (either in the supplemental material or as a URL). [Yes/No/Not Applicable] \textcolor{blue}{Yes}
   \item All the training details (e.g., data splits, hyperparameters, how they were chosen). [Yes/No/Not Applicable] \textcolor{blue}{Yes}
         \item A clear definition of the specific measure or statistics and error bars (e.g., with respect to the random seed after running experiments multiple times). [Yes/No/Not Applicable] \textcolor{blue}{Yes}
         \item A description of the computing infrastructure used. (e.g., type of GPUs, internal cluster, or cloud provider). [Yes/No/Not Applicable] \textcolor{blue}{Not Applicable}
 \end{enumerate}

 \item If you are using existing assets (e.g., code, data, models) or curating/releasing new assets, check if you include:
 \begin{enumerate}
   \item Citations of the creator If your work uses existing assets. [Yes/No/Not Applicable] \textcolor{blue}{Yes}
   \item The license information of the assets, if applicable. [Yes/No/Not Applicable] \textcolor{blue}{Not Applicable}
   \item New assets either in the supplemental material or as a URL, if applicable. [Yes/No/Not Applicable] \textcolor{blue}{Not Applicable}
   \item Information about consent from data providers/curators. [Yes/No/Not Applicable] \textcolor{blue}{Not Applicable}
   \item Discussion of sensible content if applicable, e.g., personally identifiable information or offensive content. [Yes/No/Not Applicable] \textcolor{blue}{Not Applicable}
 \end{enumerate}

 \item If you used crowdsourcing or conducted research with human subjects, check if you include:
 \begin{enumerate}
   \item The full text of instructions given to participants and screenshots. [Yes/No/Not Applicable] \textcolor{blue}{Not Applicable}
   \item Descriptions of potential participant risks, with links to Institutional Review Board (IRB) approvals if applicable. [Yes/No/Not Applicable] \textcolor{blue}{Not Applicable}
   \item The estimated hourly wage paid to participants and the total amount spent on participant compensation. [Yes/No/Not Applicable] \textcolor{blue}{Not Applicable}
 \end{enumerate}

 \end{enumerate}

\newpage
\appendix
\onecolumn
\section{PROOFS}
\label{Proof of Theorem 1}
\subsection{Proof of Proposition \ref{prop:oracle}}
For the completeness, we provide a reproduction of the proof presented in \cite{guan2019prediction},
\begin{proof}
    We first decompose the problem in eq.~(\ref{eq:GLS}) into $K$ independent problems for different classes, referred to as the problem $P_k$:
    \begin{align}
    &\min \int_{x} \mathbbm{1}_{x\in \mathcal{A}_k}\mu(x) d x,   \label{eq:GLS2}\\
    &s.t.\; \bP[x\in \mathcal{A}_k] \geq 1-\alpha.
\end{align}
Let $\mathcal{A}_k$ be the solution to problem $P_k$, then the solution to problem eq.~(\ref{eq:GLS})  is $C(x) = \{k: x \in \mathcal{A}_k\}$.

We then define $Q(\alpha,g;F)$ is the
lower $\alpha$ percentile of a real-valued function $g(x)$ under distribution $F$, i.e.,
\begin{equation}
Q(\alpha,g;F) = sup\{t: \bP_{F}(g(x)\leq t)\leq \alpha\}.
\end{equation}

Following \cite{guan2019prediction}, we regard the problem $P_k$ as a hypothesis testing problem where the null hypothesis is $H_0: x \sim f_k$ and the alternative is $H_1: x \sim \mu_k$. The optimal solution of $P_k$ is $\mathcal{A}_k$ which is actually the decision region of above hypothesis with the most powerful level $\alpha$. Therefore, by Neyman–Pearson Lemma \citep{neyman1933ix}, we can construct the likelihood ratio statistic $s(x, k;\mu) = f_k(x)\slash\mu(x)$  and have the solution $\mathcal{A}_k = \{s(x, k;\mu) \leq Q(\alpha,s_k;F_k) \}$ where the $s_k$ is the conformal score function of class $k$ and $F_k$ is the distribution of $x$ from class $k$. Hence, the solution to eq.~((\ref{eq:GLS})  is $C(x) =\{k: \bE_X[\mathbbm{1}\{s(x,k; \mu)\geq s(X,k;\mu)\}|Y=k]\geq \alpha,k=1,\ldots, K\}$.

\end{proof}
\subsection{Proof of Theorem \ref{thm:coverage}}
\begin{proof}
Here, we prove eq.~(\ref{eq:coverage}) for any given class $k$ and the test sample $x_i$. The original procedure for determining whether we should assign class $k$ to sample $x_i$ can also be described as following. First, generate two events, $\mathcal{E}_1$ and $\mathcal{E}_2$:
\begin{enumerate}
    \item  Event $\mathcal{E}1$: Training samples other than class $k$ and bootstrap copies $\mI_{other}^b$ for $b=1,\ldots, \tilde B$.
    \item Event $\mathcal{E}2$: Test samples other than $x_i$ and bootstrap copies $\mI_{test}^b$ for $b=1,\ldots, \tilde B$.
\end{enumerate}
Let $\mathcal{I}_k^b$ for $b=1,\ldots, \tilde B$ represent $\tilde B$ bootstrap copies of training class $k$ samples. We conduct our conformalized classification using only copies $b$ with $v_b=1$ for $b=1,\ldots, \tilde B$, where $v_b\sim {\rm Bernoulli}((1-\frac{1}{n_k+1})^{n_k})$. The comparison between $x_{i'}$ from class $k$ and the test sample $x_i$ is performed by aggregating prediction functions using runs $b$ with $v_b=1$ and excluding both $x_{i'}$ and $x_i$: $\hat G^{ii'}(x; \mu) = \varphi({\hat G_{k}^{b}(x;\mu): b\text{ satisfies } v_b=1, i'\notin \mathcal{I}^b_k, i\notin \mI_{te}^b})$.

We will now condition on  $\mathcal{E}_1$ and $\mathcal{E}_2$, and define $\mathcal{B}_{i}=\{b: i\notin \mathcal{I}_{te}^b\}$. The function $\hat G^{ii'}(x; \mu)$ can be rewritten as $\hat G^{ii'}(x; \mu)= \varphi({\hat G_{k}^{b}(x;\mu): b\in \mathcal{B}_{i}\text{ satisfies } v_b=1, i'\notin \mathcal{I}^b_k})$. A key observation is that this can be equivalently expressed as first sampling $B\sim {\rm Binomial}(|\mathcal{B}{i}|, (1-\frac{1}{n_k+1})^{n_k})$ and constructing the ensemble prediction function comparing $x_i$ and $x_{i'}$: $\hat G^{i'}(x; \mu) = \varphi({\hat G_{k}^{b}(x;\mu): b\text{ satisfies }i'\notin \mathcal{I}^b_k, 1\leq b\leq B})$. We have also dropped the superscript $i$ since this dependence disappears after conditioning and restricting ourselves to $\mathcal{B}_{i}$.

Interestingly, under this new equivalent characterization,  sampling $B\sim {\rm Binomial}(|\mathcal{B}_{i}|, (1-\frac{1}{n_k+1})^{n_k})$ followed by bootstrap $B$ copies of the $n_k$ class $k$ samples is equivalent to drawing $|\mathcal{B}{i}|$ bootstrap copies of the $n_k+1$ samples, which include $n_k$ samples from training class $k$ and the test sample $x_i$, and then keeping only those bootstrap samples where $x_i$ is not included. A similar equivalence was first noted by \cite{kim2020predictive} and utilized in the Jacknife+aB procedure for traditional supervised regression.

In summary, the decision rule in CSForest for whether to include label $k$ in $\hat C(x_i)$ can be equivalently expressed with the following procedure after conditioning on $\mathcal{E}_1$ and $\mathcal{E}_2$:
\begin{enumerate}
    \item Conduct $|\mathcal{B}{i}|$ bootstrap resamplings of the $n_k+1$ samples. Denote these samples as $x_1,\ldots, x_{n_k}$ (representing $n_k$ class $k$ training samples) and $x_{n_k+1}\leftarrow x_i$ as the test sample $x_i$. Let $\tilde I_b$ be the index of samples in the $b^{th}$ bootstrap.
    \item For each bootstrap, construct a random forest tree $\tilde G^b(x)$ separates $x_{\tilde I_b}$ from other samples (conditioned on).
    \item For each pair $(l, j)$ with $1\leq l, j\leq n_k+1$, construct $ \hat G^{lj}(x) =  \varphi(\{\tilde G^{b}_k(x): l,j\notin \tilde I_b\})$   and include label $k$ if and only if
    \[
    1+\sum_{j=1}^n \mathbbm{1}\{\hat G^{n_k+1,j}(x_{n_k+1})\geq \hat G^{j,n_k+1}(x_j)\}\geq (n_k+1)\alpha.
    \]
\end{enumerate}

Note that $\hat{s}^{ii'}(x,k;\mu)$ is the same as $G^{n_k+1,i'}(x_{n_k+1})$.

Define $A_{lj} = \mathbbm{1}\{\hat G^{l,j}(x_l)\geq \hat G^{j,l}(x_j)\}$ for all $1\leq l,j\leq n_k+1$. It is obvious that $A_{ii}=1$ for all $i=1,\ldots, n+1$. Define $A_{l\bullet} = \sum_{j=1}^{n_k+1} A_{lj}$ as the sum of the $l^{th}$ row from the comparison matrix A. Then, 
\[
k\notin \hat C_l(x_{n_k+1})\quad \mbox{if and only if}\qquad A_{n_k+1\bullet}\leq (n_k+1)\alpha - 1.
\]
Hence, eq.~(\ref{eq:coverage}) from Theorem \ref{thm:coverage} is equivalent to (\ref{eq:proof1}) below:
\begin{equation}
    \label{eq:proof1}
    \bP(A_{n_k+1\bullet}\leq (n_k+1)\alpha - 1|y_{n_k+1}=k)\leq 2\alpha.
\end{equation}
We now proceed to prove (\ref{eq:proof1}), which consists of two steps (1) $A_{j\bullet}$ are exchangeable with each other for $j=1,\ldots, n+1$ when $y_{n_k+1}=k$, and (2) the strange set $S(A) = \{j: A_{j\bullet}\leq (n_k+1)\alpha-1\}$ satisfies $|S(A)|\leq 2\alpha(n_k+1)$. Combining these two steps, we immediately have
\begin{align*}
    &\bP\left[A_{n_k+1\bullet}\leq (n_k+1)\alpha - 1|y_{n_k+1}=k\right] \\
   = &\bP\left[(n_k+1)\in S(A)|y_{n_k+1}=k\right]= \frac{|S(A)|}{n_k+1}\leq 2\alpha.
\end{align*}
At this stage, proofs to above two steps (1) and (2) become identical to that used in the proofs to Theorem 1 in \cite{barber2021predictive} or  Theorem 1 in \cite{kim2020predictive}.

\end{proof} 

\section{MORE DETAILS ON BASELINES}
\label{More details on baselines}
In this section, we provide more details about the baseline models. We first introduce several existing methods for constructing set-valued predictions $C(x)$, including BCOPS, CRF, DC, and ACRFrandom. 

\subsection{BCOPS, CRF, DC and ACRFrandom}

\begin{itemize}
\item  BCOPS (\underline{B}alanced \underline{C}onformalized \underline{O}ptimal \underline{P}rediction \underline{S}ets) \citep{guan2019prediction} is a semi-supervised classifier that utilizes half of the training data to train a classifier, separating observed classes from unlabeled test samples. The remaining half of the training samples is used for calibration through conformal prediction and constructing a set-valued prediction set. In contrast, BCOPS focuses on optimizing model performance on the test set, setting $\mu(x) = f_{te}(x)$, which represents the marginal density for the test data. It constructs calibrated set-valued predictions by combining empirically estimated $v_k(x)$ with the sample-splitting conformal prediction method\citep{vovk2005algorithmic}. While BCOPS excels in abnormality detection, outperforming non-test-cohort-adaptive methods, it relies on having a large set of test data, and the sample-splitting scheme results in lower data utilization efficiency\citep{guan2019prediction}.

\item CRF (\underline{C}omformalized \underline{R}andom \underline{F}orest)\citep{vovk2005algorithmic} constructs the set-valued prediction $\{k: \hat p_k(x)\geq \tau_k\}$ by including training labels $k$ achieving high estimated probability from the random forest classifier, with the cut-offs $\tau_k$ chosen based on sample-splitting conformal prediction.

\item DC  (\underline{D}ensity-set \underline{C}lassifier)\citep{hechtlinger2018cautious} constructs the set-valued prediction similarly to CRF, except for replacing the estimated probability $\hat p_k(x)$ by an estimation of the density function for class $k$ using the training data.

\item ACRFrandom:  We refer to the adaptive-coverage classification approach using random forest proposed in \cite{romano2020classification} as ACRFrandom (Adaptive-coverage CRF with randomization) where the randomization is introduced via an additional uniform random variable $U$ for tie-breaking. ACRFrandom constructs the prediction set $\hat C(x)$ by including labels with large estimated probabilities such that the total probability is greater than the upper-level quantile of the empirical distribution of $\{E_i\}_{i\in \mI_{cal}}\cup \{\infty\}$ where $\mI_{cal}$ is the calibration set in sample-splitting conformal prediction and  $E_i$ is the sum of estimated probabilities for all labels proceeding that for the true label. 

\end{itemize}

\subsection{ACRF and ACRFrandom}
We further introduce the baseline ACRFrandom and its de-randomized version ACRF.
\subsubsection*{ACRFrandom}
In the original proposal of \cite{romano2020classification}, the authors assume that training and test data to have the same distribution. ACRFrandom defines a function $\mathcal{S}$ with input $x$, $u \in [0,1]$, the conditional probability $\pi_y=P(Y=y|X=x)$, and the threshold $\tau$. Define
\begin{equation}
\label{eq:ACRF-S}
\mathcal{S}(x,u;\pi,\tau)=\left\{
\begin{array}{cl}
y\;\text{indices of the}\; L(x;\pi,\tau)-1\; \text{largest}\; \pi_y(x), & \text{if } u < V(x;\pi,\tau) \\
y\;\text{indices of the}\; L(x;\pi,\tau)\; \text{largest}\; \pi_y(x), & \text{otherwise}
\end{array}
\right.
\end{equation}

where

\begin{equation}
\label{eq:ACRF-S-V}
V(x;\pi,\tau)=\frac{\sum_{c=1}^{L(x;\pi,\tau)}\pi_{(c)}(x)-\tau}{\pi_{(L(x;\pi,\tau))}(x)}
\end{equation} 

\begin{equation}
\label{eq:ACRF-S-L}
L(x;\pi,\tau)=\min\{c \in \{1,\cdots,C\}: \pi_{(1)}(x)+\pi_{(2)}(x)+\cdots+\pi_{(c)}(x) \geq \tau\}.
\end{equation}

and $\pi_{(i)}(x)$ is the $ith$ largest conditional probability. 

Further, ACRFrandom defines the generalized inverse quantile conformity score function $E$,

\begin{equation}
\label{eq:ACRF-S-E}
E(x,y,u;\hat{\pi})=\min\{\tau \in [0,1]: y\in \mathcal{S}(x,u;\hat{\pi},\tau)\}.
\end{equation}

And the empirical distribution is

\begin{equation}
\label{eq:ACRF-e}
V(x,y;E)=\frac{1}{|\mI_{cal}|+1}\sum_{i \in \mI_{cal}}\delta_{E_i}+\frac{1}{|\mI_{cal}|+1}\delta_{\infty}
\end{equation}

where $E_i$ is constructed at the minimum $\tau$ for the calibration sample $i$ such that $y_i$ is included in $S(x_i, u_i;\hat\pi, \tau)$ and $\delta_i$ denotes a point mass at $i$. The final prediction set is constructed as  $\hat C(x) = \mathcal{S}(x,u;\pi, \hat\tau_{\alpha})$,
where $\hat\tau_{\alpha}$ is the upper level $\alpha$ quantile of the empirical distribution $\{E_i\}_{i\in \mI_{cal}}\cup \{\infty\}$.

\subsubsection*{ACRF}

We consider a derandomized version of ACRF without the uniform variable $U$ in our experiment. For ACRF,  we define



\begin{equation}
\label{eq:ACRF}
\mathcal{S}(x;\pi,\tau)=\{y\;indices\;of\;the\;L(x;\pi,\tau)\;largest\;\pi_y(x)\},
\end{equation}
where
\begin{equation}
\label{eq:ACRF-L}
L(x;\pi,\tau)=\min\{c \in \{1,\cdots,C\}: \pi_{(1)}(x)+\pi_{(2)}(x)+\cdots+\pi_{(c)}(x) > \tau\}.
\end{equation}
We can similarly define a score function $E$,
\begin{equation}
\label{eq:ACRF-E}
E(x,y;\hat{\pi})=\min\{\tau \in [0,1]: y\in \mathcal{S}(x;\hat{\pi},\tau)\},
\end{equation}
and construct $E_i$ as the the minimum $\tau$ for the calibration sample $i$ such that $y_i$ is included in $S(x_i;\hat\pi, \tau)$. Same as in ACRFrandom, ACRF constructs as  $\hat C(x) = \mathcal{S}(x;\pi, \hat\tau_{\alpha})$,
where $\hat\tau_{\alpha}$ is the upper level $\alpha$ quantile of the empirical distribution $\{E_i\}_{i\in \mI_{cal}}\cup \{\infty\}$. Algorithm \ref{alg:ACRF} shows the details of ACRF.

 \begin{algorithm}[H]
  \caption{Implementation of ACRF}
   \label{alg:ACRF}
 
\SetKwInOut{Input}{Input}
\SetKwInOut{Output}{Output}
\setcounter{AlgoLine}{0}
\Input{Training Data $\{z_i\coloneqq(x_i, y_i)^n_{i=1},i\in \mI_{tr}\}$,  Test Data $\{(x_i)^m_{i=1},i\in \mI_{te}\}$.} 
\Output{Prediction sets $\hat C_i(x_i)$ for $i\in \mI_{te}$.}

Randomly split the training data into 2 subsets, the training set $\mI^1_{tr}$, the calibration training set $\mI^2_{tr}$.

Train random forest model $\mathcal{B}$ on all samples in $\mI^1_{tr}$: $\hat{\pi}_1 \leftarrow \mathcal{B}({(X_i, Y_i)}_{i\in\mI^1_{tr}})$.

Predict on $\mI_{tr}^2$: $\hat{\pi}_2 \leftarrow \mathcal{B}({(X_i)}_{i\in\mI_2})$ and $\mI_{te}$: $\hat{\pi}_{te} \leftarrow \mathcal{B}({(X_i)}_{i\in\mI_{te}})$.

Construct $\{E_i\}_{i\in \mI^{2}_{tr}}$ treating $\mI^{2}_{tr}$ as the calibration set.

Compute the level $(1-\alpha)$ quantile of the empirical distribution $\{E_i\}_{i\in \mI^{2}_{tr}}\cup\{\infty\}$.

Use the function $\mathcal{S}$ defined in eq.~({\ref{eq:ACRF}}) to construct the prediction set at $x_i \in \mI_{te}$ as 
$\hat{C}_i(x_i)=\mathcal{S}(X_i;\hat{\pi}^i_{te},\hat{\tau}_{\alpha})$.
 
\end{algorithm}

\subsection{ACRFshift}

Finally, we introduce ACRFshift, another baseline that explicitly accounts for distributional changes under covariate shift model. ACRFshift combines the covariate shift comformal prediction  \citep{tibshirani2019conformal} with ACRF. This has not been  discussed in previous work, so we give details about how this is done in our paper. We  split also the test samples into two sets $\mI_{te}^1$, $\mI_{te}^2$. Suppose that we now construct the prediction set for test samples in $\mI_{te}^2$.

Instead of finding $\tau$ with eq.~(\ref{eq:ACRF-E}) and (\ref{eq:ACRF-e}), we consider the following weighted calibration. The weighted function is $\gamma_{x}(x) = \frac{r(x)}{r(x)+\sum_{z_i\in \mI_2} r(x_i)}$ and $r(x) = \frac{\bP[W=1|x]}{\bP[W=0|x]}$ is the conditional probability of being generated from the test data ($W=1$ means from the test data, and $W=0$ represent from the training data), learned from the classifier separating the test data $\mI_{te}^1$ from the training data $\mI^1_{tr}$. Instead of consider $\hat\tau_{\alpha}$ as the level $(1-\alpha)$ quantile of the empirical distribution  $\{E_i\}_{i\in \mI^{2}_{tr}}\cup\{\infty\}$, for any $x\in \mI_{te}^2$, we consider the level $(1-\alpha)$ quantile of the weighted distribution below:
\[
V_w(x,y;E)=\sum_{i \in \mI_{tr}^2}\gamma_x(x_i)\delta_{E_i}+\gamma_x(x)\delta_{\infty}.
\]
Similarly, whether to include the random variable $U$ in ACRF will result in two versions: ACRFshiftrandom and ACRFshift.

Unlike CSForest, BCOPS, and even CRF, which naturally considers samples with $\emptyset$ as the ones not close to inlier classes, and thus, outliers, ACRF and ACRFshift both consider the conditional probability of $y|x$ and do not have such a feature encoded in their constructions. Hence, we adopt the rule where we reject a sample when $r_x(x)$ is very large compared to others with $\sum_{i\in I_{tr}^2} r_x(x_i)< \tau$. (Recall that $r_x(x_i)=\frac{1}{n+1}$ without covariate shift.)

\subsection{Randomized ACRF/ACRFshift vs. derandomized ACRF/ACRFshift}
\label{Randomized ACRF vs.derandomized ACRF}
In this section, we further demonstrate the difference between the randomized version and derandomized version to support our claim that removing the random variable $U$ from ACRFrandom and ACRFshift helps achieve a desirable type II error.

Table \ref{table:randomized and derandomized} shows the type I error and type II error of ACRF/ACRFrandom and ACRFshift/ACRFshiftrandom (referred to as derandomized and randomized versions of ACRF/ACRFshift in Table \ref{table:randomized and derandomized}). We observe that models with randomness and those without randomness achieve comparable type I errors when there is not shift in the inlier labels, while models with randomness tend to have worse type II errors and the final prediction set $\hat{C}$ is more likely to contain multiple labels. One potential explanation for this is that randomization can help with the control of  conditional coverage  and may lead to increased high type II error due to this more ambitious goal. Supporting this, the randomized version for both ACRF and ACRFshift controls the type I error while the derandomized version, especially ACRF, shows inflated type I error under the label shift model.

\begin{table}[H]
\caption{Randomized version vs. derandomized version: Achieved Type I and Type II errors at $\alpha = 0.05$ with outlier components and no additional label shift among inlier digits and achieved Type I and Type II errors at $\alpha = 0.05$ with label shift among inlier digits but no outlier digits.  }
\label{table:randomized and derandomized}
\begin{center}
\begin{small}
\begin{sc}
\begin{tabular}{lccccr}
\toprule
\multirow{2}{*}{Version}&\multirow{2}*{Method}& \multicolumn{2}{c}{no additional label shift}&\multicolumn{2}{c}{additional label shift}\\
\cmidrule{3-6} 
&&Type I&Type II&Type I&Type II\\
\midrule
\multirow{2}*{Randomized}&ACRF&0.049$\pm$0.007&  0.702$\pm$0.017&{0.025$\pm$0.007}&{0.884$\pm$0.014}\\
&ACRFshift &0.053$\pm$0.007&0.681$\pm$0.014&{0.055$\pm$0.013}& 0.828$\pm$0.015\\ 
\midrule
\multirow{2}*{derandomized}&ACRF&0.047$\pm$0.006   &   0.431$\pm$0.003 &0.171$\pm$0.024    &   0.313$\pm$0.067\\
&ACRFshift&0.036$\pm$0.009    &   0.439$\pm$0.009 &0.080$\pm$0.026    &  0.630$\pm$0.127\\ 
\bottomrule
\end{tabular}
\end{sc}
\end{small}
\end{center}
\end{table}

\section{TRAINING DETAILS}
\label{Training Details}
In this section, we provide a detailed description of how we constructed datasets satisfying GLS and label shift using MNIST, CIFAR-10, and FashionMNIST for experimentation.

\textbf{Q1 \textbf{outliers w/o shift}.} For the datasets MNIST, CIFAR-10, and FashionMNIST, each consisting of 10 categories, we sampled 500 samples from each class (categories 0-5) to create a training set of 2500 samples. From the remaining samples in categories 0-5, we randomly selected 500 samples from each class, and similarly, we randomly selected 500 samples from each class in categories 6-9. These 5000 samples formed the test set. Notably, in the test set, categories 6-9 represent outliers that never appeared in the training set, while categories 0-5 are inlier samples. 

\textbf{Q2 \textbf{shift w/o outliers}.} For the Label Shift setup, we sampled 500 samples from each class (categories 0-5) from MNIST, CIFAR-10, and FashionMNIST to create a training set of 3000 samples. From the remaining samples in categories 0-5, we randomly selected 100 samples from each class, and similarly, we randomly selected 500 samples from each class in categories 6-9. These 3000 samples formed the test set.

It is important to emphasize that for MNIST, CIFAR-10, and FashionMNIST, both CSForest and the baseline methods utilized representations extracted by a pre-trained ResNet-18 model as inputs, rather than the original images.

\section{ADDITIONAL EXPERIMENTS RESULTS}
\label{More Results}
In the following section, we present additional experimental results to further substantiate the conclusions made in Section \ref{sec:experiments}.

\subsection{More Datastes}
\label{more data}

To further illustrate the effectiveness of CSForest under different tasks, we additionally include experiments on two new datasets: a cyber/network intrusion dataset from the KDD data competition and a chest X-ray dataset from the medical domain. Results in Tabel \ref{tab:add results} shows CSForest achieves low type I errors and minimizes type II errors in both datasets, demonstrating its superior capability for outlier detection compared to the baseline.

\begin{table}[H]
\caption{The achieved type I and type II errors at $\alpha = 0.05$ under no additional distribution shift but with outliers in the test set across 10 repetitions. For Network Intrusion, the test set includes additional 15 intrusion types as outliers; for Chest X-ray, the test set includes lung abnormalities caused by viruses as outliers.}
\centering
\begin{tabular}{lccccccc}
\toprule
\multirow{2}{*}{Method} & \multicolumn{2}{c}{Network Intrusion}        & \multicolumn{2}{c}{Chest X-ray}                \\ \cmidrule(lr){2-3} \cmidrule(lr){4-5}  & Type I Error&Type II Error &Type I Error&Type II Error\\ 
\midrule
CSForest&  0.048 $\pm$ 0.013   &    \textbf{9.524e-5 $\pm$ 0.000}&  0.056 $\pm$ 0.003 & \textbf{0.566 $\pm$ 0.002}  \\ 
BCOPS&0.047 $\pm$ 0.014   &   5.905e-4 $\pm$ 0.000&     0.059 $\pm$ 0.005&  0.576 $\pm$ 0.004 \\ 
DC  &  0.049 $\pm$ 0.010 &   0.261$\pm$ 0.049&     0.062 $\pm$ 0.010  &    0.728 $\pm$ 0.013  \\ 
CRF   &  0.030 $\pm$ 0.004 & 0.467$\pm$ 0.104 &     \textbf{0.039 $\pm$ 0.033}   &    0.793 $\pm$ 0.178  \\ 
ACRF& { 0.000 $\pm$ 0.000}   &  0.857$\pm$ 0.000&    {0.067 $\pm$ 0.046}    &   0.922 $\pm$ 0.005\\
ACRFshift & 0.001 $\pm$ 0.001   & 0.019 $\pm$ 0.004&    {0.039 $\pm$ 0.049}   & 0.885 $\pm$ 0.057\\
\bottomrule
\end{tabular}
\label{tab:add results}
\end{table}

\subsection{Per-class Quality Evaluation}
\label{Per-class quality evaluation}

\begin{figure}[H]
  \centering
  \subfigure[Per-class quality evaluation with outliers but no additional label shift among inlier digits, where the outliers are defined as $R = \{6,7,8,9\}$. ]{
    \includegraphics[width=0.46\textwidth]{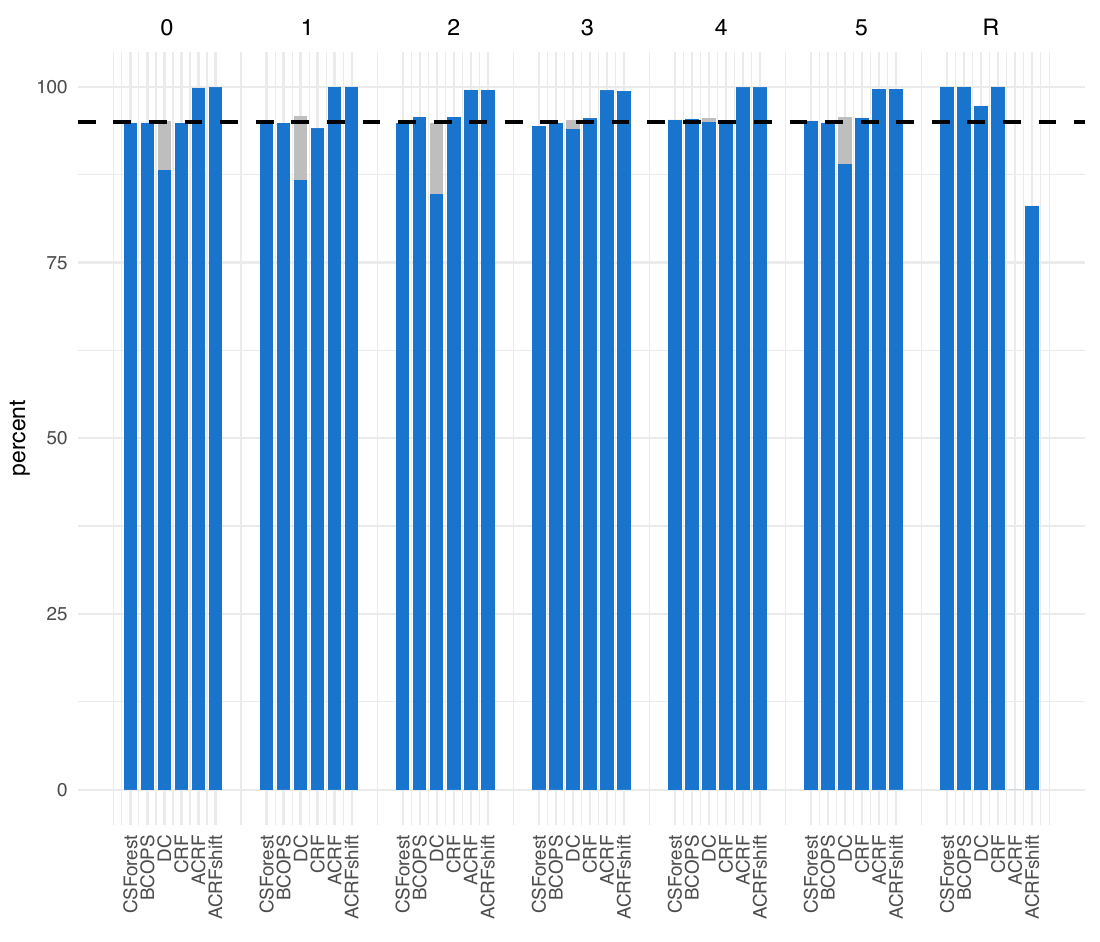}
    \label{fig:CoverageRateBarCIFAR}
  }
  \subfigure[Per-class (class 0-5) quality evaluation with additional label shift among inlier digits but no outliers.]{
    \includegraphics[width=0.44\textwidth]{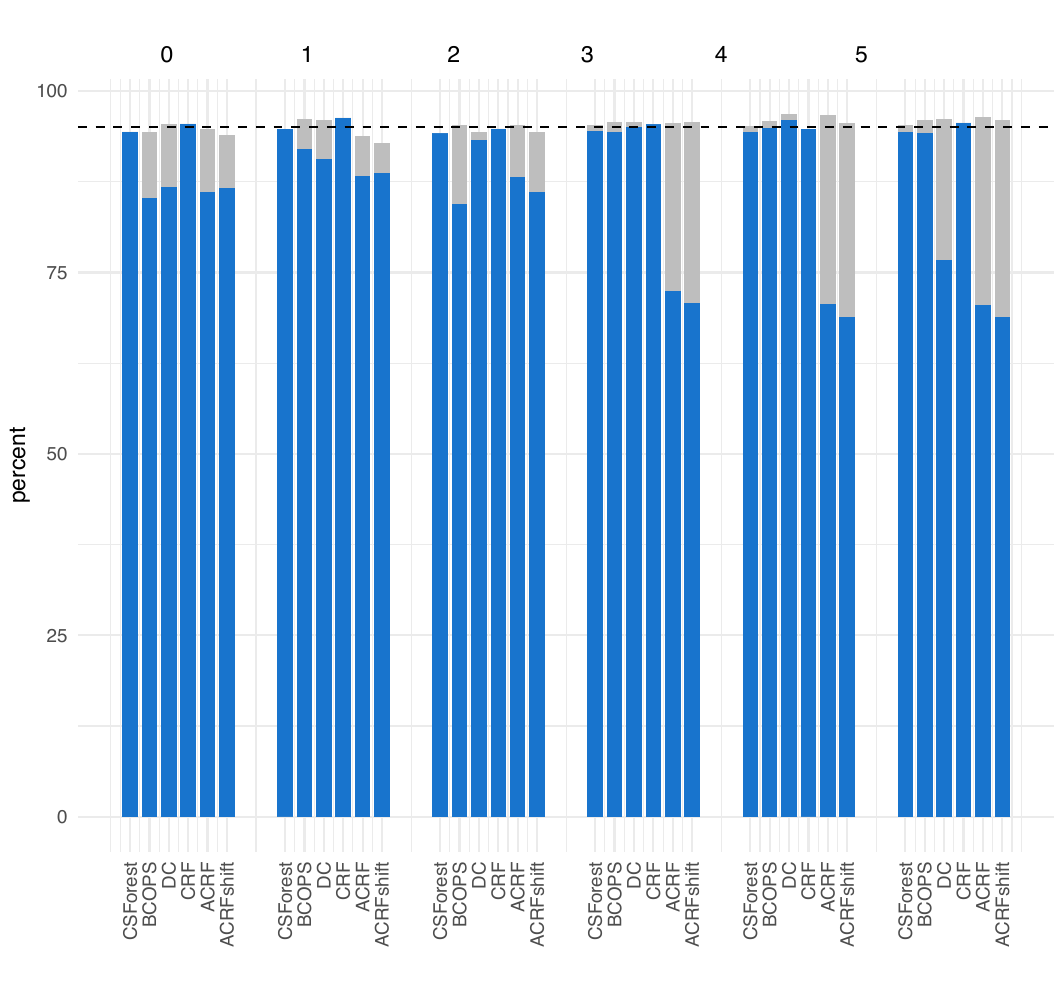}
    \label{fig:CoverageRateBarCIFAR_No}
  }
  \caption{Per-class quality evaluation on CIFAR-10. Figure \ref{fig:per_class_cifar} is grouped by the actual labels in the testing data and colored based on if a prediction set contains only the correct label (blue) or more than the correct label (gray).  The horizontal dash line refers to the coverage level of 95\%.}
  \label{fig:per_class_cifar}
\end{figure}

\begin{figure}[H]
  \centering
  \subfigure[Per-class quality evaluation with outliers but no additional label shift among inlier digits, where the outliers are defined as $R = \{6,7,8,9\}$.]{
    \includegraphics[width=0.46\textwidth]{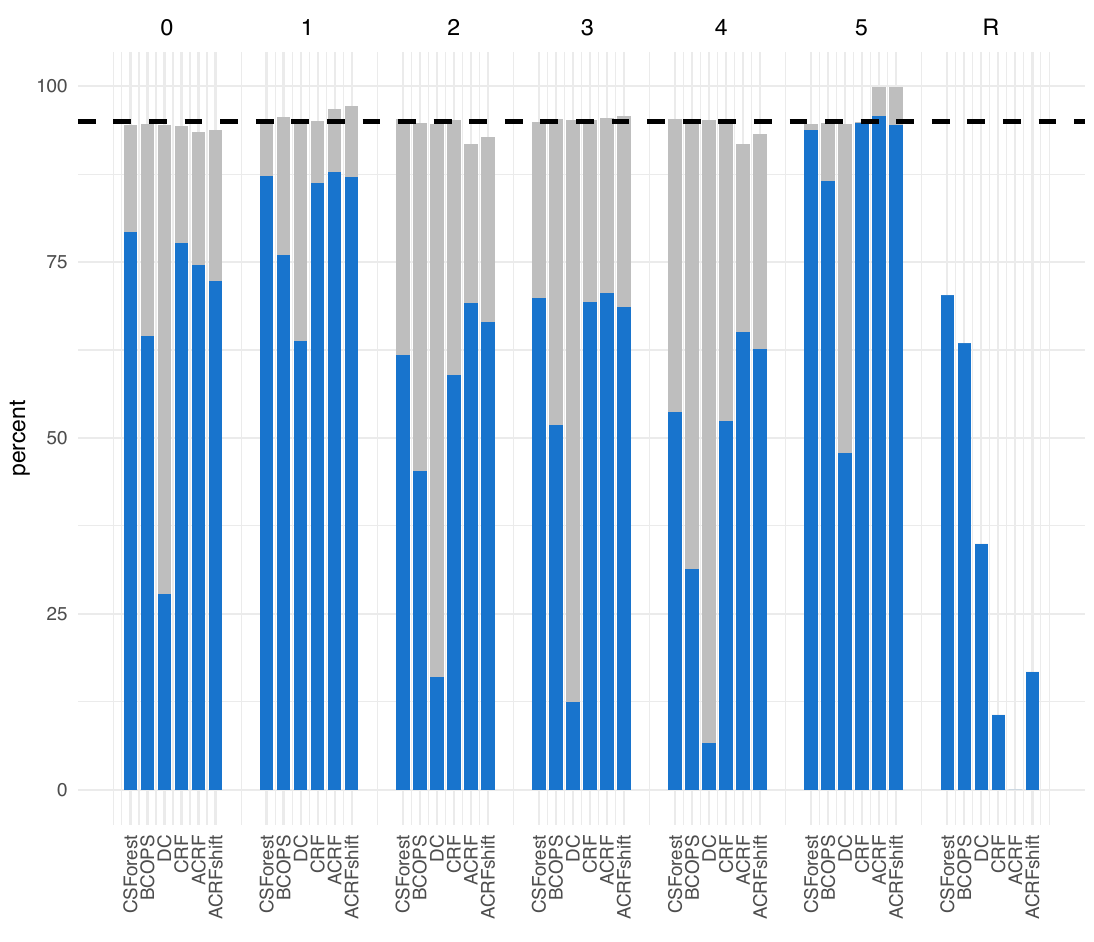}
    \label{fig:CoverageRateBarFashionMNISTI}
  }
  \subfigure[Per-class (class 0-5) quality evaluation with additional label shift among inlier digits but no outliers.]{
    \includegraphics[width=0.44\textwidth]{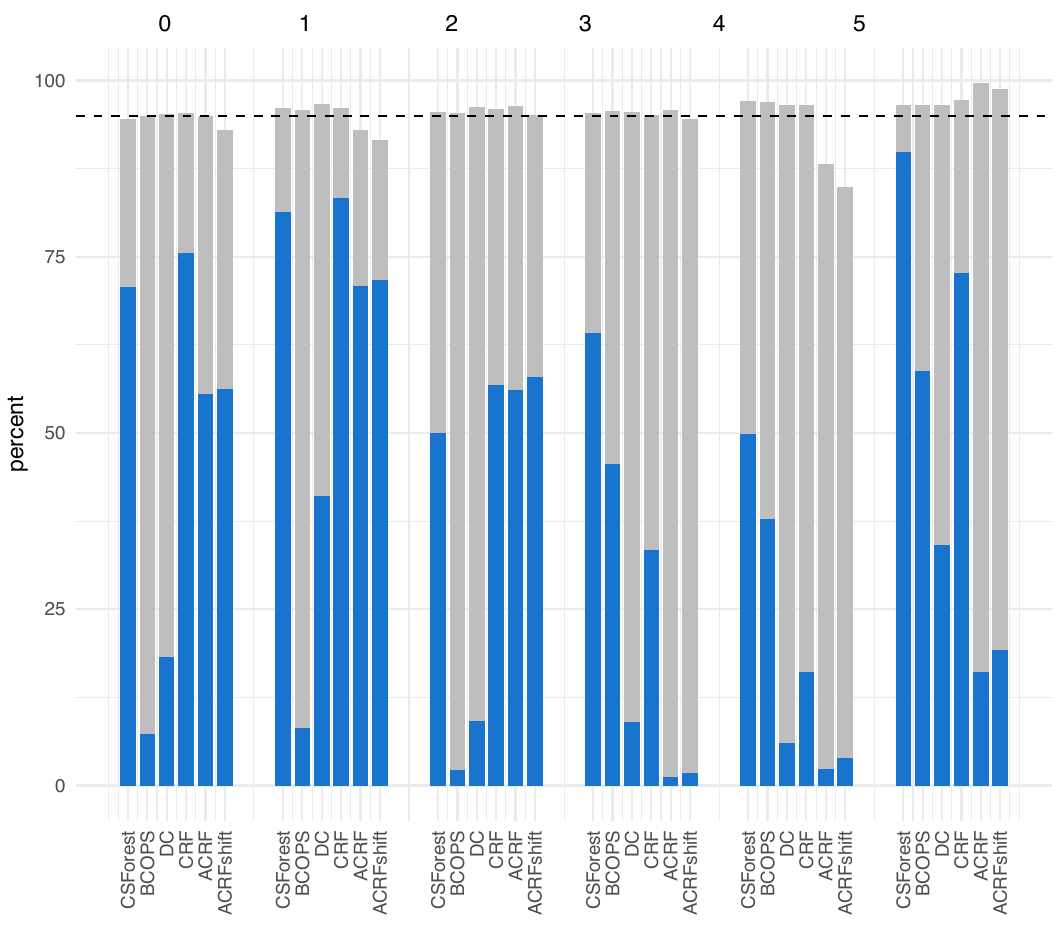}
    \label{fig:CoverageRateBarFashionMNISTI_No}
  }
  \caption{Per-class quality evaluation on FashionMNIST. Figure \ref{fig:per_class_fashionmnist} is grouped by the actual labels in the testing data and colored based on if a prediction set contains only the correct label (blue) or more than the correct label (gray).  The horizontal dash line refers to the coverage level of 95\%.}
  \label{fig:per_class_fashionmnist}
\end{figure}

Figures \ref{fig:per_class_cifar} and \ref{fig:per_class_fashionmnist} provide a detailed breakdown of the predictions made by all methods for each class in the CIFAR-10 and FashionMNIST datasets. Consistent with the results observed in MNIST, we find that CSForest is the top-performing method for outlier detection, and it avoids over-predicting by generating prediction sets that predominantly contain only the correct labels for each class.

\subsection{Average Length of the Prediction Set $\hat{C}$}
\label{Average length of the prediction set}

We observe that across all datasets, whether in the setting with outliers and no additional label shift or without outliers and with additional label shift, CSForest consistently achieves the smallest average prediction set interval length. This indicates that CSForest's predictions do not contain a significant amount of redundant information, aligning with our previous observations of CSForest containing more ``only correct labe" content for each class and exhibiting lower type II error.

\begin{figure}[H]
  \centering
  \subfigure[Per-class quality evaluation of all methods with outlier components but no additional label shift among inlier digits. ]{
    \includegraphics[width=0.4\linewidth]{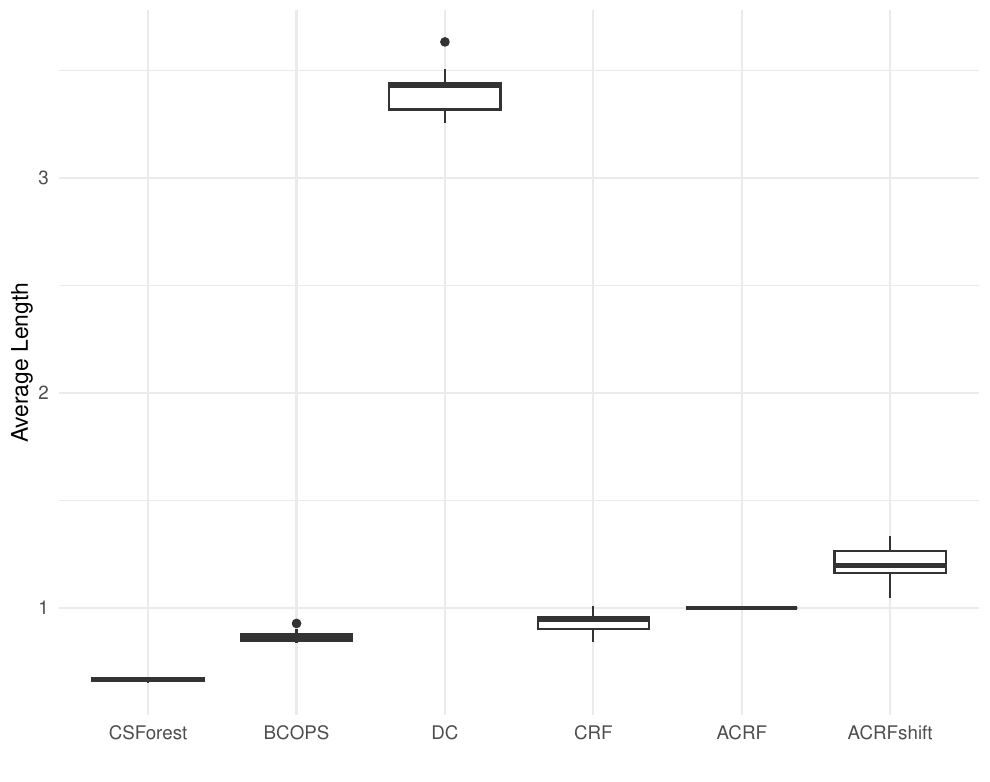}
    \label{fig:len_MNISTI}
  }
  \subfigure[Average length of the prediction set $\hat{C}$ of all methods with additional label shift among inlier digits but no outlier components.]{
    \includegraphics[width=0.4\linewidth]{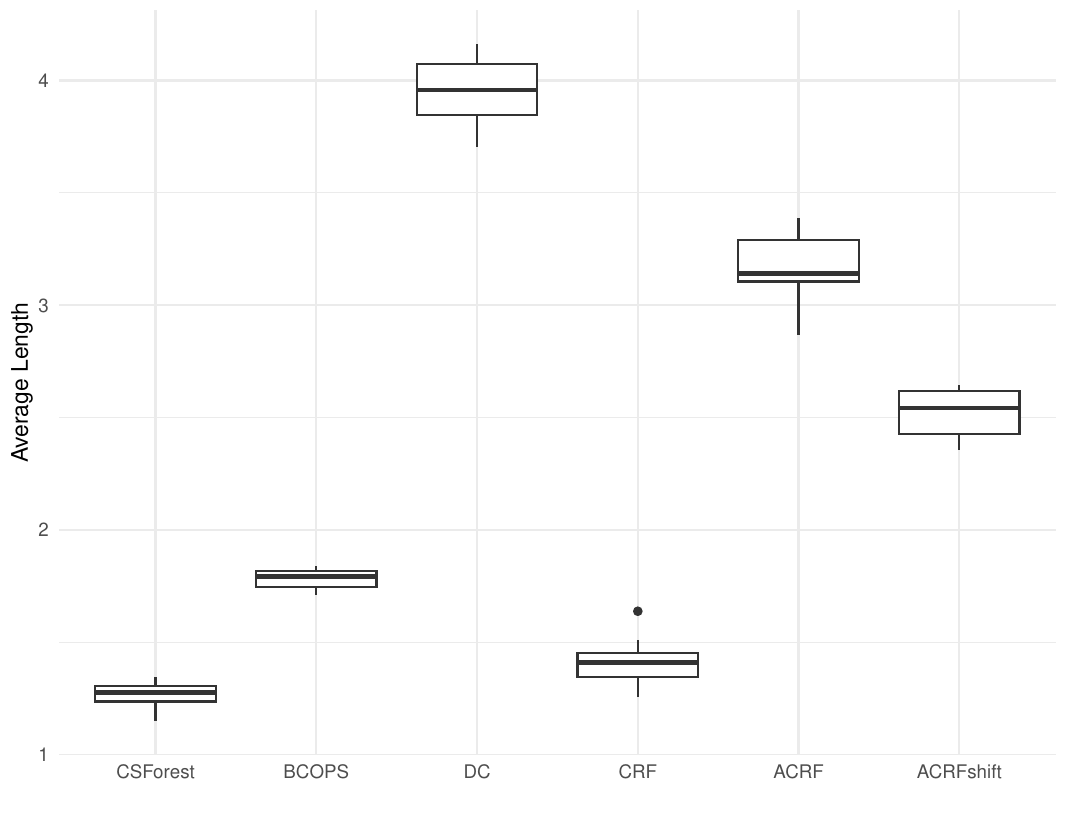}
    \label{fig:len_MNISTI_No}
  }
  \caption{Average length of the prediction set $\hat{C}$ on MNIST. For MNIST, in both settings, CSForest achieves the smallest average prediction set interval length, which aligns with the high ``only correct labels" content demonstrated in Figure \ref{fig:per_class_mnist} for CSForest across all classes. This conclusion is further supported by the lower type II error exhibited by CSForest in Table \ref{tab:all results}.}
  \label{fig:len_mnist}
\end{figure}

To further illustrate that CSForest's prediction set $\hat{C}$ results in more accurate label predictions (i.e., predominantly containing only the correct labels) compared to other methods, Figure \ref{fig:len_mnist}, \ref{fig:len_cifar} and \ref{fig:len_fashionmnist} visualize the average interval length of prediction sets $\hat{C}$ for all methods.

\begin{figure}[H]
  \centering
  \subfigure[Per-class quality evaluation of all methods with outlier components but no additional label shift among inlier digits. ]{
    \includegraphics[width=0.4\textwidth]{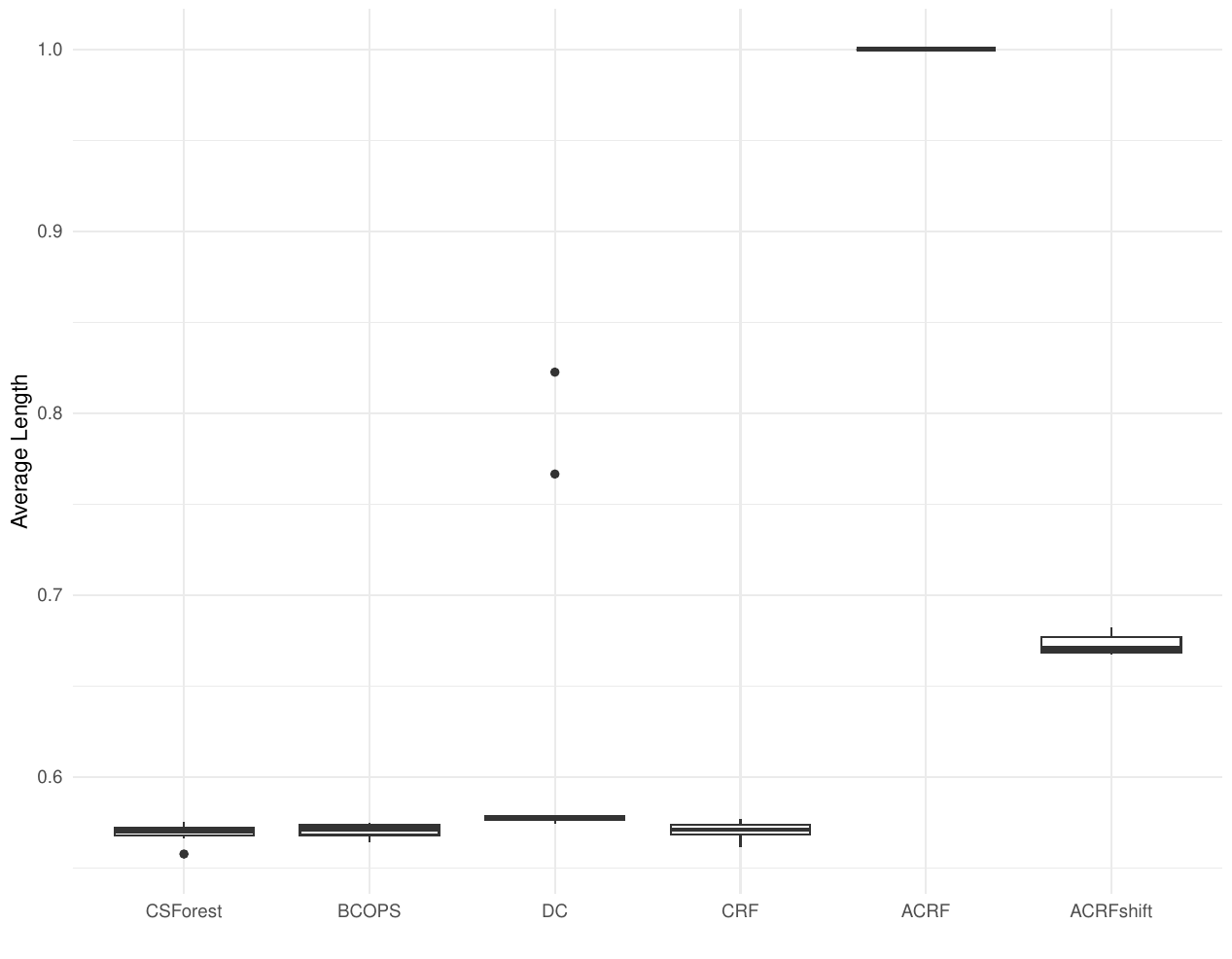}
    \label{fig:len_cifar}
  }
  \subfigure[Average length of the prediction set $\hat{C}$ of all methods with additional label shift among inlier digits but no outlier components.]{
    \includegraphics[width=0.4\textwidth]{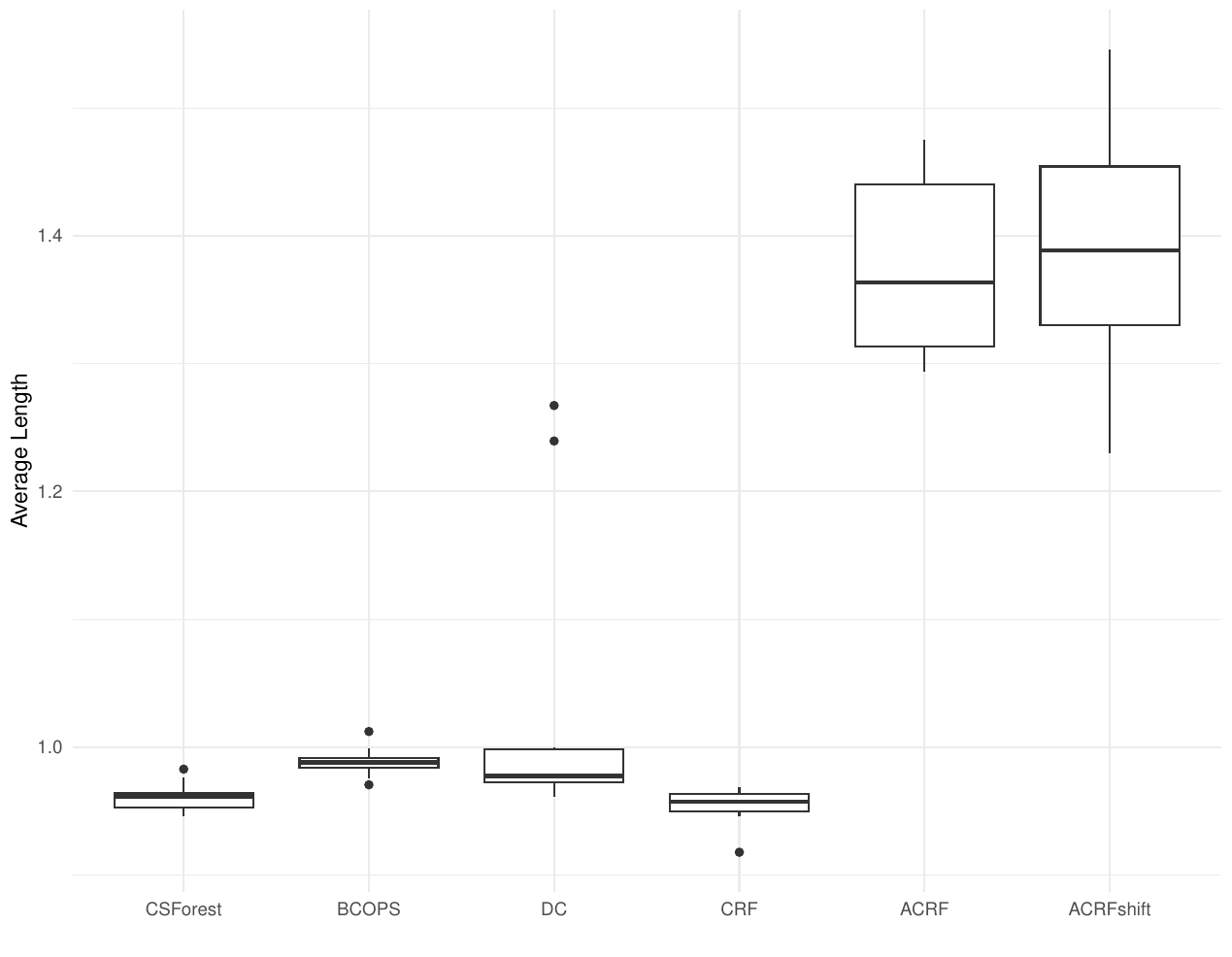}
    \label{fig:len_cifar_No}
  }
  \caption{Average length of the prediction set $\hat{C}$ on CIFAR-10. For CIFAR-10, in both settings, CSForest achieves the smallest average prediction set interval length, which aligns with the high ``only correct labels" content demonstrated in Figure \ref{fig:per_class_cifar} for CSForest across all classes. This conclusion is further supported by the lower type II error exhibited by CSForest in Table \ref{tab:all results}.}
  \label{fig:len_cifar}
\end{figure}

\begin{figure}[H]
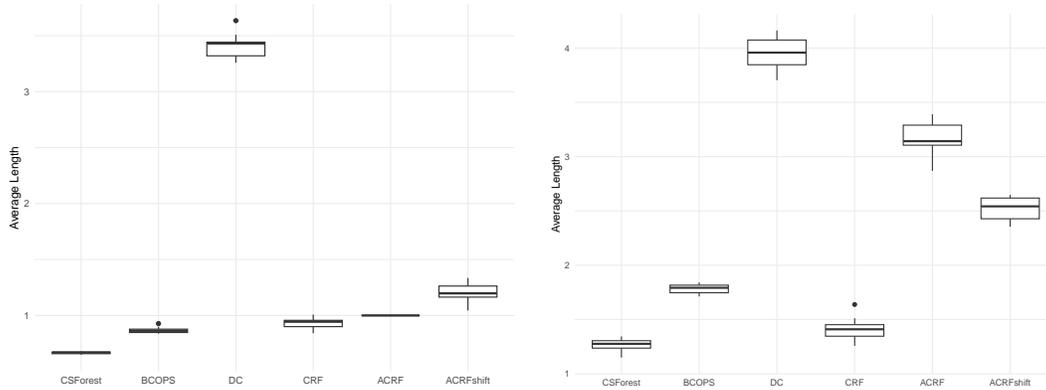

  \centering
  \subfigure[Per-class quality evaluation of all methods with outlier components but no additional label shift among inlier digits. ]{
    \includegraphics[width=0.4\textwidth]{png/length_mnist.pdf}
    \label{fig:len_fashionmnist}
  }
  \subfigure[Average length of the prediction set $\hat{C}$ of all methods with additional label shift among inlier digits but no outlier components.]{
    \includegraphics[width=0.4\textwidth]{png/length_mnist_nooutlier.pdf}
    \label{fig:len_fashionmnist_No}
  }
  \caption{Average length of the prediction set $\hat{C}$ on FashionMNIST. For FashionMNIST, in both settings, CSForest achieves the smallest average prediction set interval length, which aligns with the high ``only correct labels" content demonstrated in Figure \ref{fig:per_class_fashionmnist} for CSForest across all classes. This conclusion is further supported by the lower type II error exhibited by CSForest in Table \ref{tab:all results}.}
  \label{fig:len_fashionmnist}
\end{figure}

\subsection{CSForest's Performance with Varying Sample Size}
\label{app:sample szie}
In this section, we present the type II (inlier and outlier) error curves for all methods on CIFAR-10 and FashionMNIST as sample sizes vary. Consistent with the experimental results on MNIST, CSForest demonstrates superior outlier detection capabilities relative to the baseline as sample sizes change, and it also maintains lower inlier type II errors.


\begin{figure}[H]
  \centering
  \includegraphics[width=0.8\textwidth, height=0.35\textheight]{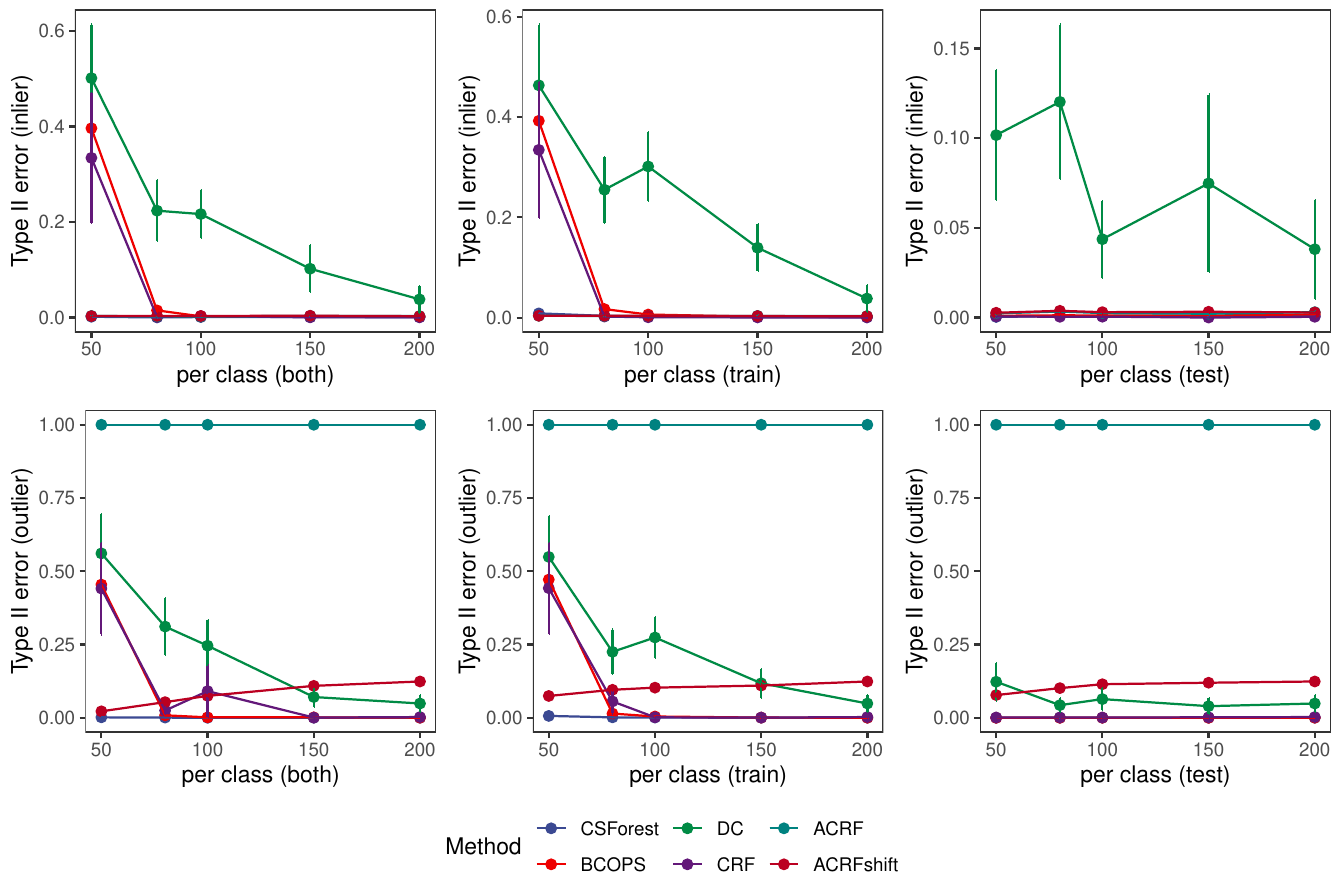}
  \caption{The type II error for inliers and outliers obtained under different sample sizes on CIFAR-10. Figure \ref{fig:CIFARvaringsize} illustrates that, compared to baselines, CSForest efficiently detects outliers while maintaining lower inlier type II errors across varying sample sizes. }
  \label{fig:CIFARvaringsize}
\end{figure}

\begin{figure}[H]
  \centering
  \includegraphics[width=0.8\textwidth, height=0.35\textheight]{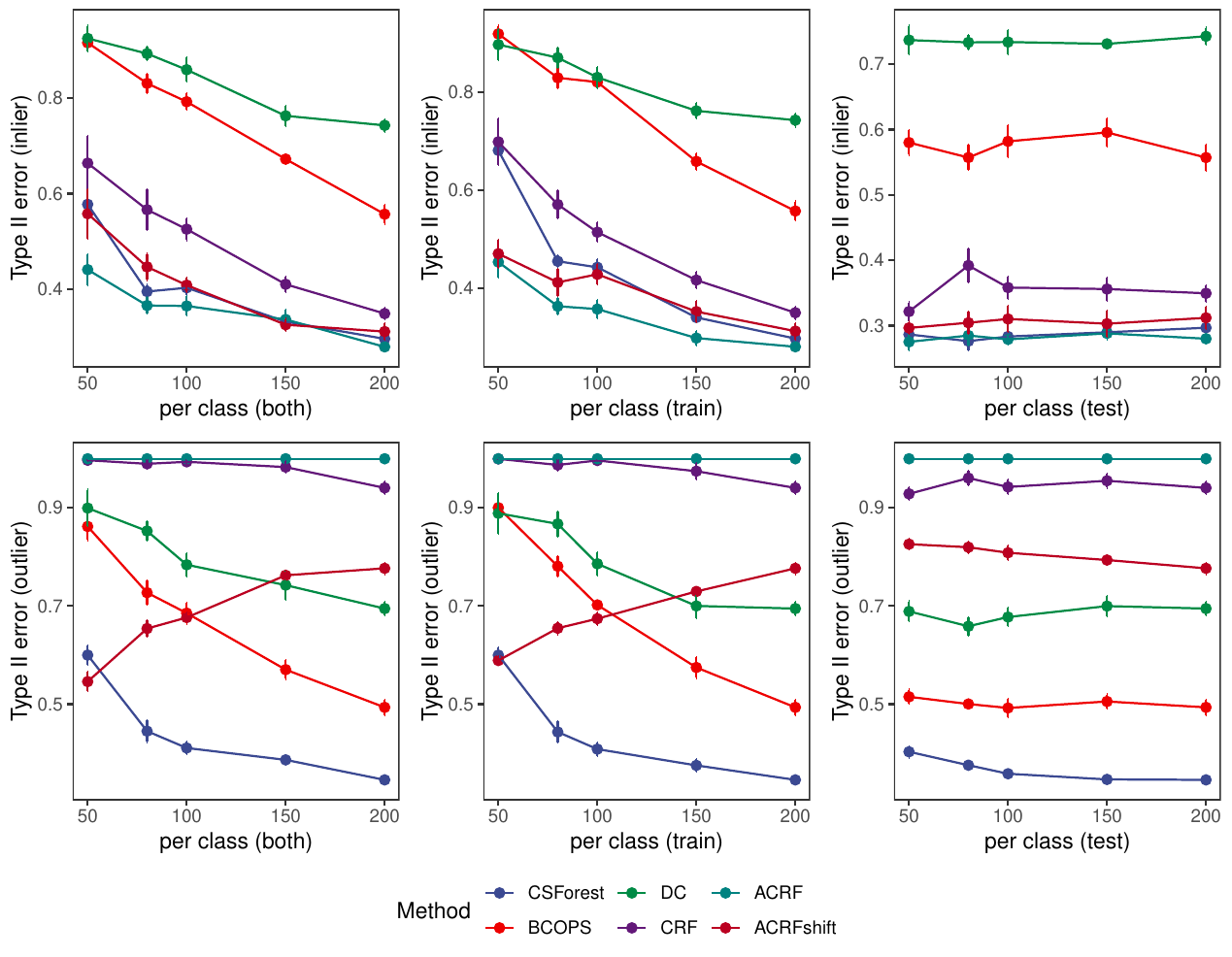}
  \caption{The type II error for inliers and outliers obtained under different sample sizes on FashionMNIST. Figure \ref{fig:FashionMNISTvaringsize} illustrates that, compared to baselines, CSForest efficiently detects outliers while maintaining lower inlier type II errors across varying sample sizes. }
  \label{fig:FashionMNISTvaringsize}
\end{figure}

\section{DISCUSSION ON $w$}
\label{Discussion on w}
To avoid oversampling, we impose constraints on $\tilde\mI_{other}$, where the sample size for other classes is constrained to $\min(\lceil n_{te}w \rceil, n - n_k)$. That is, we weigh the training samples through sampling, and we will cap the influence of training samples if even taking all training samples becomes insufficient. So the threshold of satisfies $\lceil n_{te}w \rceil=n - n_k$. As a result, CSForest will yield the same Type I and Type II error once  $\omega$ exceeds a certain threshold depending on the data.

Assuming a training dataset with $K$ classes and $T$ samples per class, and a test dataset with $K'$ classes and $T'$ samples per class. We have:

\begin{equation}
\label{eq:threshold w}
\begin{aligned}
n_{te}\omega + 1 &\geq n - n_k \geq n_{te}\omega \\
\implies K'T'\omega + 1 &\geq KT - T \geq K'T'\omega \\
\implies \frac{(K-1)T}{K'T'} &\geq \omega \geq \frac{(K-1)T - 1}{K'T'}
\end{aligned}
\end{equation}

Based on eq.~(\ref{eq:threshold w}), we have the following conclusions:

\begin{enumerate}
    \item If $\omega$ exceeds the threshold $\frac{{(K-1)T}}{{K'T'}}$, increasing $\omega$ will have no effect on CSForest.
    \item If the test sample size $n_{te}=K'T' \gg KT=n_{tr}$ where $n_{tr}$ is the training sample size, $\omega \to 1$. In this case, greater than 1 will achieve the same errors as $\omega=1$.
\end{enumerate}

In this section, we experiment with different choices of weights $w$ from small $0$ to large (exceeding the weight threshold) as opposed to fixing $w=1$ at the default value. Although small $w$ can sometimes lead to improved outlier detection and large $w$ can sometimes improve the inlier classification, $w=1$ tend to provides a good tradeoff between these two objectives on the three real data sets considered.

\subsection{MNIST}

For the MNIST data, we consider $K=6, T=200, K'=10, T'=50$ and we can get the threshold $1.998<w<2.000$. For the MNIST dataset, we indeed observed in Table \ref{tab:Errw_mnist} that once $w \geq 2$, the type I error and type II error of CSForest no longer change. 

\begin{table}
\caption{Achieved Type I and Type II errors at 
$\alpha=0.05$ and $\omega \geq 1$ on MNIST. We observed that when $\omega \geq 2$, CSForest achieves the same Type I and Type II error as 
$\omega \geq 2$.}
\begin{center}
\begin{small}
\begin{sc}
\begin{tabular}{cccc}
\toprule
\textbf{$\omega$}&Type I Error&Type II Error (inlier)& Type II Error (outlier)\\
\midrule
0&   0.057$\pm$0.018 &    0.251$\pm$0.033 &   0.314$\pm$0.068 \\ 
LOG&   0.053$\pm$0.016 &    0.224$\pm$0.031 &   0.315$\pm$0.061  \\ 
1&     0.058$\pm$0.014 &    0.119$\pm$0.018 &   0.346$\pm$0.065\\ 
1.5&     0.055$\pm$0.016   &    0.106$\pm$0.014    & 0.349$\pm$0.068 \\ 

2& 0.056$\pm$0.018   &    0.099$\pm$0.016 &  0.373$\pm$0.072  \\ 
5&   0.056$\pm$0.018   &    0.099$\pm$0.016 &  0.373$\pm$0.072 \\ 
10&  0.056$\pm$0.018   &    0.099$\pm$0.016 &  0.373$\pm$0.072   \\
100&  0.056$\pm$0.018   &    0.099$\pm$0.016 &  0.373$\pm$0.072   \\
\bottomrule
\end{tabular}
\label{tab:Errw_mnist}
\end{sc}
\end{small}
\end{center}
\end{table}

\subsection{CIFAR-10}

For CIFAR-10, we consider $K=6, T=200, K'=10, T'=50$ and get the threshold $1.998<w<2.000$. Similarly to MNIST, for CIFAR-10, once $w$ exceeds the threshold of 2, the performance of CSForest remains unchanged.

\begin{table}
\caption{Achieved Type I and Type II errors at 
$\alpha=0.05$ and $\omega \geq 1$ on CIFAR-10. We observed that when $\omega \geq 2$, CSForest achieves the same Type I and Type II error as 
$\omega \geq 2$.}
\begin{center}
\begin{small}
\begin{sc}
\begin{tabular}{cccc}
\toprule
\textbf{$\omega$}&Type I Error&Type II Error (inlier)& Type II Error (outlier)\\
\midrule
0&   0.045$\pm$0.015 &    0.001$\pm$0.001 &   0.000$\pm$0.000 \\ 
LOG&   0.043$\pm$0.016 &    0.000$\pm$0.001 &   0.000$\pm$0.000  \\ 
1&     0.043$\pm$0.016 &    0.000$\pm$0.001 &   0.000$\pm$0.000\\ 
1.5&    0.043$\pm$0.014 &    0.000$\pm$0.001 &   0.000$\pm$0.000 \\ 

2& 0.043$\pm$0.016 &    0.000$\pm$0.001 &   0.000$\pm$0.000  \\ 
5&   0.043$\pm$0.016 &    0.000$\pm$0.001 &   0.000$\pm$0.000 \\ 
10&  0.043$\pm$0.016 &    0.000$\pm$0.001 &   0.000$\pm$0.000   \\
100&  0.043$\pm$0.016 &    0.000$\pm$0.001 &   0.000$\pm$0.000   \\
\bottomrule
\end{tabular}
\label{tab:Errw_cifrar10}
\end{sc}
\end{small}
\end{center}
\end{table}

\subsection{FashionMNIST}

For the CIFAR-10 data, we consider $K=6, T=200, K'=10, T'=50$ and get the threshold  $1.998<w<2.000$. For FashionMNIST as well, the performance of CSForest remains constant once $w$ surpasses the threshold of 2.

\begin{table}
\caption{Achieved Type I and Type II errors at 
$\alpha=0.05$ and $\omega \geq 1$ on FashionMNIST. We observed that when $\omega \geq 2$, CSForest achieves the same Type I and Type II error as 
$\omega \geq 2$.}
\begin{center}
\begin{small}
\begin{sc}
\begin{tabular}{cccc}
\toprule
\textbf{$\omega$}&Type I Error&Type II Error (inlier)& Type II Error (outlier)\\
\midrule
0&   0.049$\pm$0.011 &    0.421$\pm$0.001 &   0.406$\pm$0.045 \\ 
LOG&   0.048$\pm$0.012 &    0.391$\pm$0.047 &   0.399$\pm$0.043  \\ 
1&     0.046$\pm$0.012 &    0.287$\pm$0.028 &   0.403$\pm$0.037\\ 
1.5&    0.047$\pm$0.010 &    0.272$\pm$0.029 &   0.408$\pm$0.038 \\ 

2& 0.048$\pm$0.012 &    0.262$\pm$0.032 &   0.410$\pm$0.034  \\ 
5&   0.048$\pm$0.012 &    0.262$\pm$0.032 &   0.410$\pm$0.034 \\ 
10&  0.048$\pm$0.012 &    0.262$\pm$0.032 &   0.410$\pm$0.034  \\
100&  0.048$\pm$0.012 &    0.262$\pm$0.032 &   0.410$\pm$0.034  \\
\bottomrule
\end{tabular}
\label{tab:Errw_fashionmnist}
\end{sc}
\end{small}
\end{center}
\end{table}

\vfill

\end{document}